\documentclass[10pt,journal,compsoc]{IEEEtran}
\ifCLASSOPTIONcompsoc
  \usepackage[nocompress]{cite}
\else
  \usepackage{cite}
\fi

\ifCLASSINFOpdf
  \usepackage[pdftex]{graphicx}
\else
  \usepackage[dvips]{graphicx}
\fi
\usepackage{amsmath}
\usepackage{algorithm,algorithmic}
\usepackage{url}
\usepackage{comment,color}
\usepackage{subfigure,times}
\usepackage{amssymb}

\newtheorem{proposition}{Proposition}
\newcommand{\pf}{\noindent{\bf Proof:~}}

\DeclareMathAlphabet{\mathsfbf}{OT1}{cmss}{sbc}{n}

\newcommand{\EE}{\mathbb{E}} 
\newcommand{\VV}{\mathbb{V}} 
\newcommand{\PP}{\mathbb{P}} 

\newcommand{\ee}{{\rm e}}
\newcommand{\dd}{{\rm\,d}} 

\newcommand{\erf}{{\rm erf}} 

\newcommand{\dv}{{\bf d}}

\newcommand{\qv}{{\bf q}}

\newcommand{\xv}{{\bf x}}

\newcommand{\zerov}{{\bf 0}}

\newcommand{\Am}{{\bf A}}

\newcommand{\Id}{{\bf I}}

\newcommand{\Mm}{{\bf M}}

\newcommand{\Ec}{{\cal E}}

\newcommand{\Gc}{{\cal G}}

\newcommand{\Ic}{{\cal I}}

\newcommand{\Kc}{{\cal K}}
\newcommand{\Lc}{{\cal L}}

\newcommand{\Nc}{{\cal N}}

\newcommand{\Sc}{{\cal S}}
  
\newcommand{\Uc}{{\cal U}}
\newcommand{\Wc}{{\cal W}}

\newcommand{\deltav}{\boldsymbol{\delta}}

\newcommand{\rhov}{\boldsymbol{\rho}}

\newcommand{\thetav}{\boldsymbol{\theta}}

\newcommand{\sigmav}{\boldsymbol{\sigma}}
\newcommand{\piv}{\boldsymbol{\pi}}

\newcommand{\Gammam}{\boldsymbol{\Gamma}}

\newcommand{\Deltam}{\boldsymbol{\Delta}}
\newcommand{\Sigmam}{\boldsymbol{\Sigma}}

\def\Tran{^\mathsf{T}}

\def\ben{\begin{enumerate}}
\def\beq{\begin{equation}}
\def\beqa{\begin{eqnarray}}
\def\bit{\begin{itemize}}
\def\een{\end{enumerate}}
\def\eeq{\end{equation}}
\def\eeqa{\end{eqnarray}}
\def\eit{\end{itemize}}

\def\non{\nonumber\\}

\newtheorem{prop}{Proposition}

\begin{document}

\title{Ranking a Set of Objects using Heterogeneous Workers: QUITE an Easy Problem}

\author{Alessandro Nordio,~\IEEEmembership{Member, IEEE},
      Alberto Tarable,
      Emilio Leonardi,~\IEEEmembership{Senior Member, IEEE}%
  \IEEEcompsocitemizethanks{\IEEEcompsocthanksitem A.~Nordio and A. Tarable are with CNR-IEIIT (Institute of Electronics,
  Telecommunications and Information Engineering of the National Research Council of Italy),
  Italy, email: {firstname.lastname}@ieiit.cnr.it.%
  \IEEEcompsocthanksitem E. Leonardi is with DET, Politecnico di Torino, Torino, Italy and is also associate researcher with CNR-IEIIT.}}%

%
%

%

\IEEEtitleabstractindextext{%
\begin{abstract}
    We focus on the problem of ranking $N$ objects starting from a set
    of noisy pairwise comparisons provided by a crowd of unequal
    workers, each worker being characterized by a specific degree of
    reliability, which reflects her ability to rank pairs of objects.
    More specifically, we assume that objects are endowed with
    intrinsic qualities and that the probability with which an object
    is preferred to another depends both on the difference between the
    qualities of the two competitors and on the reliability of the worker.
    We propose QUITE, a non-adaptive ranking algorithm that 
    jointly estimates workers'  reliabilities and 
    qualities of objects.  Performance of QUITE is compared in
    different scenarios against previously proposed algorithms.  Finally,
    we show how QUITE can be naturally made adaptive.
\end{abstract}

}

\maketitle

\IEEEdisplaynontitleabstractindextext

%
\IEEEpeerreviewmaketitle

\IEEEraisesectionheading{\section{Introduction and related work}}

\IEEEPARstart{T}{his} paper focuses on the problem of establishing a reliable ranking
among several objects, starting from a set of noisy human evaluations.
Such problem emerges in several computer-science contexts, e.g.  when
web pages are ranked by search engines, when hotels and restaurants
are ranked by applications like Tripadvisor,  or when products are
ranked by on-line sellers~\cite{sponsoredads, NIPS2006_3079}.

Often, a ranking algorithm receives as input a set of
noisy preferences between pairs of objects and infers an estimated
order relation among them. Comparisons are sometimes made by human
workers, whose behavior cannot deterministically predicted. Indeed,
outcomes of comparisons depend on how objects are ``perceived'' by
human workers,  rather than on their intrinsic quality.

This specific problem has attracted a significant bulk of attention in
the last few years \cite{shah2017simple, d2019ranking,
  falahatgar2017maxing,noiTNSE,cinesi},  and a set of stochastic laws
has been proposed to represent the behavior of human workers, such as
the very popular
Bradley-Terry-Luce~\cite{plackett1975analysis,bradley1952rank,luce2012individual}
and Thurstone~\cite{thurstone1927method} models.  Most of them are
based on the assumption that an intrinsic quality can be associated to
every object and that the probability that object $i$ is preferred to
object $j$ depends on the associated qualities $q_i$ and $q_j$.  The
vast majority of previous works \cite{shah2017simple, d2019ranking,
  falahatgar2017maxing,noiTNSE,szorenyi2015online,falahatgar2017maximum,falahatgar2018limits}
assumes workers to be homogeneous, i.e., to behave exactly according
to the same law.  For such simplified scenario several ranking
algorithms have been proposed in \cite{shah2017simple, d2019ranking,
  falahatgar2017maxing,noiTNSE} and its asymptotic performance has
thoroughly been analysed, typically within the
$(\varepsilon, \delta)$-PAC framework~\cite{szorenyi2015online,
  falahatgar2017maximum, falahatgar2018limits}.

In particular, by strengthening and generalizing previous
results~(\cite{shah2017simple, d2019ranking,
  falahatgar2017maxing,szorenyi2015online,falahatgar2017maximum,falahatgar2018limits}),
in~\cite{noiTNSE}, we have recently proposed a class of non-adaptive
ranking algorithms that rely on a least-squares (LS) optimization
criterion for the estimation of qualities.  In the scenario of
homogeneous workers with known reliability, such LS algorithms exploit
the structure of a graph $\Gc$ of cardinality $N$, whose edges are in
one-to-one correspondence with the evaluated object pairs. The LS
algorithms in~\cite{noiTNSE} are shown to be asymptotically optimal
(i.e., they require $O(\frac{N}{\epsilon^2}\log \frac{N}{\delta})$
comparisons to meet $(\varepsilon, \delta)$-PAC constraints), as long as
the graph edges are properly selected.
	Also, they operate by receiving in input the set of estimated distances
	between object pairs and by
	returning the quality estimates  as
	well as the estimated ranking. 

        Only recently, the assumption that all workers obey the same
        law has been loosen.  For example, in \cite{cinesi}, ranking
        algorithms receive as input a set of pairwise preferences
        expressed by heterogeneous users, each one obeying either a
        BTL or a Thurstone model and characterized by a different
        reliability. To the best of our knowledge only~\cite{cinesi}
        addresses this specific problem and proposes a ranking
        algorithm based on an approximate Maximum-Likelihood (ML)
        estimation of worker reliabilities and object qualities. The
        work in~\cite{cinesi} provides also a theoretical analysis of
        the algorithm convergence, from which, however, asymptotic
        properties of their algorithm can hardly be obtained.  The
        very recent work \cite{pmlr-v151-wu22f} focuses on a different
        scenario where no intrinsic qualities can be associated to
        objects, and the probability with which object $i$ is
        preferred to $j$ by worker $w$ only depends on $w$ and on the
        true ranking between $i$ and $j$.  For such scenario,
        \cite{pmlr-v151-wu22f} proposes an algorithm for ranking
        aggregation, whose performance is analytically evaluated.
	
	These are the major contributions of the present work.
\begin{itemize}
\item We propose an algorithm called QUITE, which can be successfully
  employed under a rather general class of worker behavior models,
  e.g., the generalized Thurstone model considered in~\cite{cinesi} as
  well as a generalized version of the Bradley-Terry-Luce model.  Our
  approach resorts to the graph-based LS method proposed
  in~\cite{noiTNSE} for the scenario of homogeneous workers.  At least
  in its simpler form, QUITE is amenable to a theoretical asymptotical
  analysis, which shows that, under mild conditions, it is
  asymptotically optimal (i.e., it requires
  $O(N/\varepsilon^2)\log (N/\delta)$ comparisons to comply
  $(\varepsilon, \delta)$-PAC requirements).
 \item We derive a Bayesian Cram\'er-Rao lower bound (BCRB) for
the mean-square error achievable by any estimator technique of
quality differences and/or worker reliabilities.
\item We test the performance of QUITE against the BCRB,  and compare it with the
  algorithm in \cite{cinesi}.
  \item We  extend QUITE to work in a multistage fashion, where 
  	the assignment of object pairs to workers  is made in several stages, by exploiting  previous partial estimates of  workers' reliabilities and qualities of objects.
In such a case, we propose a simple recipe for assigning workers to object
  pairs, on the base of estimations performed at previous stages.
\end{itemize}	

The rest of the manuscript is organized as follows:
Section~\ref{sec:model} describes the system model and discusses
possible approaches for obtaining a ranking.  Section~\ref{sec:BCRB}
derives the expression of the BCRB on the variance of the
quality-reliability estimates. Section~\ref{sec:ranking_algorithms}
provides a review of ranking algorithms available in the literature
and introduces our proposed joint quality-reliability estimation
algorithm, i.e., QUITE. Section~\ref{sec:2stage} deals with a
multistage version of QUITE and gives also a heuristic rule of
assignment of objects pairs to workers, on the basis of previous
output of the algorithm.  In Section~\ref{sec:results}, we provide
numerical results showing the performance of QUITE in several
scenarios.
Finally,  in Section~\ref{sec:conclusion}, we draw the conclusions of our work.

\subsection{Mathematical notation}
Boldface uppercase and lowercase letters denote matrices and vectors,
respectively. The $(i, j)$-th entry of matrix $\Am$ is denoted by
$[\Am]_{i,j}$ and its transpose is denoted by $\Am\Tran$. Calligraphic
letters denote sets or graphs. The symbols $\EE_\xv[\cdot]$, $\VV_\xv[\cdot]$ and
$\nabla_\xv$ are, respectively, the expectation operator, the variance operator and the
gradient w.r.t.  vector $\xv$. The probability of an event
$x$ is denoted by $\PP(x)$ while $f_x(x)$ indicates the probability density function of
the random variable $x$. A uniform probability distribution with support $[a,b]$ is denoted by $\Uc[a,b]$
whereas the Gaussian distribution with mean $\mu$ and variance $\sigma$ is denoted by $\Nc(\mu,\sigma)$. 

\section{System description\label{sec:model} }
We consider the problem of ranking a set of $N$ objects having unknown
intrinsic qualities $q_1, \dots, q_N$, where $q_i \in \mathbb{R}$.  A
ranking is a relationship between the elements in the set,  such that
an object ranked higher is considered ``better'', ``superior'', or
``preferred'' to an object ranked lower.  For example, the qualities
$q_i$ induce a true ranking, $r$, in which $r(i) \prec r(j)$ (i.e.,
object $i$ is better than object $j$) iff $q_i > q_j$. In other words,
the true ranking corresponds to a permutation, $\piv$, of the integers
$\{1,\ldots,N\}$ defined by sorting object qualities,  such that
$q_{\pi_1} > q_{\pi_2} > \dots > q_{\pi_N} $\footnote{We suppose
  ties happen with probability zero.}.

A ranking algorithm provides an estimated ranking, $\widehat{r}$, by
resorting to some information obtained by comparing the objects. Such
estimated ranking corresponds to a permutation $\widehat{\piv}$ of the
integers $\{1,\dots,N\}$, with the meaning that object
$\widehat{\pi}_1$ is ranked the best, and object $\widehat{\pi}_N$
the worst.  We say that $\widehat{r}$ is an $\epsilon$-quality ranking
if $\widehat{r}(i) \prec \widehat{r}(j)$ whenever $q_i > q_j
+\epsilon$. Moreover a ranking algorithm is ($\varepsilon$, $\delta$)-PAC
(probably approximately correct)~\cite{szorenyi2015online,
  falahatgar2017maximum, falahatgar2018limits} if it returns an
$\epsilon$-quality ranking with a probability larger than $1-\delta$.

A ranking algorithm processes information obtained, e.g., through a
set of observations or comparisons made by a pool of workers.  In particular, we assume that a set $\Ec$ of distinct object pairs, having cardinality
$E=|\Ec|$, is given to a pool of $K$ workers for evaluation; for each
assigned pair, a worker gives a binary answer, indicating the object
in the pair he/she ranks higher.  Since workers are not fully
reliable, the same pair is given to several of them. Workers'
answers can be considered as independent random variables and modeled
as described in the following.

Let $(i_e, j_e)\in \Ec$ be the $e$-th object pair, $e = 1,\dots,E$ and
denote by $d_e = q_{i_e} - q_{j_e}$ the quality difference between
objects in pair $e$.  Also, let $\Ec_k \subseteq \{1,\dots,E\}$ be the
subset of pairs assigned to worker $k$, $k= 1,\dots,K$ and $\Kc_e
\subseteq \{1,\dots,K\}$ be the subset of workers assigned to pair
$e$, $e= 1,\dots,E$.  The set of binary outcomes of the evaluation is
denoted by $\Wc = \{w_{e,k}\}$, where $w_{e,k}\in\{0,1\}$ represents
the outcome of the evaluation provided by worker $k$ on object pair
$e\in\Ec_k$.  In particular $w_{e,k}=0$ if worker $k$ prefers object
$i_e$ when evaluating pair $e$ and $w_{e,k}=1$ otherwise.  The
preferred object is chosen in accordance to quality {\em as perceived}
by worker $k$ and defined as $q_{i_e,k}^{\rm perceived}=
q_{i_e}+\eta_{k,e,i_e}$ where $\eta_{k,e,i_e}$ is a random evaluation noise, modeled
as independent for each object. In practice the worker outputs
$w_{e,k}=0$ if $q_{i_e,k}^{\rm perceived} > q_{j_e,k}^{\rm perceived}$.
In order to characterize workers' evaluations,  we  assign to worker $k$ an unknown reliability
parameter $\rho_k \in \mathbb{R}^+$, $k=1,\ldots,K$. Workers with
large $\rho_k$ provide highly reliable answers, i.e., they are very
sensitive to small quality differences between objects. Instead,
answers provided by workers characterized by small $\rho_k$ tend to be
less correlated to the actual qualities of the objects being evaluated.

Summarizing, given $d_e$ and $\rho_k$, the conditional probability of the
outcome $w_{e,k}=0$ is defined as
\begin{equation}
  \PP(w_{e,k}=0|\rho_k,d_e) \triangleq F(\rho_k,d_e)
  \label{eq:P_w0}
\end{equation}
where $F(\rho, d)$ is an increasing function of $d$ satisfying
$F(\rho,0) = 1/2$ and $F(\rho,-d) = 1-F(\rho,d)$.  In
the following, we will consider two well-known evaluation models~\cite{bradley1952rank,thurstone1927method}:
\begin{itemize}
\item \textbf{Thurstone model}, where $\eta_k$ is a
   Gaussian random variable with standard deviation
  $1/\sqrt{2\rho_k^{2}}$.   Therefore
  \begin{equation}
    F(\rho_k,d_e) = \frac1{2} \left[1 + \erf \left(\frac{\rho_k d_e}{\sqrt{2}}\right) \right]
    \label{eq:F_Thurstone}
  \end{equation}
\item \textbf{BTL model}, where $\eta_k$ is a Gumbel random
  variable with scale parameter $1/\rho_k$. Since the difference
  of two independent Gumbel rv's has a logistic distribution, we
  obtain
  \begin{equation}
    F(\rho_k,d_e) = \frac{\ee^{\rho_k d_e}}{1 + \ee^{\rho_k d_e}}
    \label{eq:F_BTL}
  \end{equation}

\end{itemize}
Since, for both models, $F(\rho_k,d_e)$ is a function of the
product $\rho_k d_e$ with a slight abuse of notation, in the
following we will write $F(\rho_k d_e)$ instead of
$F(\rho_k,d_e)$. Also, since $F(-\rho_k d_e) = 1-F(\rho_k d_e)$, we
can write in both cases
\begin{equation}
\PP(w_{e,k}|\rho_k,d_e)= F\left((1-2w_{e,k})\rho_k d_e\right).
\label{eq:pw}
\end{equation}


\section{Bayesian Cram\'er-Rao bound\label{sec:BCRB}}
In this section, we derive the Bayesian Cram\'er-Rao bound (BCRB) on
any algorithm that jointly estimates object qualities and worker
reliabilities. The BCRB allows to compute a lower bound to the
variance of any estimator of the unknown system parameters, when we
assume to have access to prior knowledge.

To do so, we first denote by $\qv=[q_1,\ldots,
  q_N]\Tran$, $\dv=[d_1,\ldots, d_E]\Tran$ and
$\rhov=[\rho_1,\ldots,\rho_K]\Tran$ the vectors of object qualities,
quality differences and worker reliabilities, respectively.  We assume
that $\rhov$ and $\qv$ are random vectors with i.i.d. entries whose
a-priori distributions are
\begin{equation}
  f_{\rhov}(\rhov) = \prod_{k=1}^K f_{\rho}(\rho_k),\qquad \mbox{ and }\qquad f_{\qv}(\qv) = \prod_{i=1}^Nf_q(q_i)\,,\label{eq:p_rho_p_q}
\end{equation}
respectively, where $f_{\rho}(\rho)$ and $f_q(q)$ are the a-priori distributions of a generic worker reliability and of an object quality, respectively. We then observe that  $\qv$ and $\dv$ satisfy the linear
relationship
\begin{equation}
  \label{eq:d_versus_q} \dv = \Gammam\Tran \qv
\end{equation}
where $\Gammam$ is an $N \times E$ matrix whose $e$-th column,
$e = 1,\dots,E$, has a 1 in row $i_e$, a -1 in row $j_e$, and 0
elsewhere.  Let $\Gc$ be the undirected graph whose $N$ nodes are in
one-to-one correspondence with objects and whose $E$ edges are in
one-to-one correspondence with the object pairs in $\Ec$. Then, $\Gammam$
can be seen as a (directed) incidence matrix of $\Gc$. The joint a-priori distribution of the distances, $f_{\dv}(\dv)$, and
the corresponding marginals, $f_{d}(d_e)$, $e = 1,\dots,E$, follow
trivially from~\eqref{eq:d_versus_q}.
 
The maximum a posteriori (MAP) joint estimate of quality distances and worker reliabilities
can be written as
\begin{eqnarray}
  && \hspace{-6ex}\{\widehat{\rhov}, \widehat{\dv}\} \non
  &=&\hspace{-2ex} \arg\max_{\rhov, \dv} \log \PP\left(\rhov,\dv | \Wc\right) \non
  &=&\hspace{-2ex} \arg\max_{\rhov, \dv} \log \left[\PP\left(\Wc | \rhov,\dv\right) f_{\rhov}(\rhov) f_{\dv}(\dv)\right]\non
  &=& \hspace{-2ex}\arg\max_{\rhov, \dv}\left( \sum_{k=1}^K \sum_{e\in \Ec_k}\log F(x_{e,k})  {+}  \log f_{\rhov}(\rhov) {+}  \log f_{\dv}(\dv)  \right)\non
  &=& \hspace{-2ex}\arg\max_{\rhov, \dv} \left( \sum_{e =1}^E \sum_{k\in \Kc_e} \log F(x_{e,k}) {+}  \log f_{\rhov}(\rhov) {+}  \log f_{\dv}(\dv)  \right)\nonumber \\
      \label{eq:MAP}
\end{eqnarray}
where we defined $x_{e,k}\triangleq(1-2w_{e,k})\rho_kd_e$, we have  exploited~\eqref{eq:pw} and assumed that the evaluation outcomes are independent, so that
\[\PP\left(\Wc | \rhov,\dv\right)= \prod_{k=1}^K \prod_{e\in \Ec_k}F(x_{e,k}) =   \prod_{e =1}^E \prod_{k\in \Kc_e}F(x_{e,k})\,.\]

Let $\thetav = [\qv\Tran,\rhov\Tran]\Tran$ be the length-$(N+K)$
vector of unknown parameters.  The BCRB allows to lower-bound the
mean-square error (MSE) between $\thetav$ and its estimate
$\widehat{\thetav}$ achieved by MAP estimation and, a fortiori, by any
other conceivable algorithm. The covariance matrix $\Sigmam$ of such
estimate is given by \beq \Sigmam = \EE_{\Wc, \thetav} \left[
  \left(\widehat{\thetav} - \thetav \right) \left(\widehat{\thetav} -
  \thetav \right)\Tran \right]. \eeq
where we recall that $\Wc$ is the set of random workers' answers.

The BCRB states that $\Sigmam
\succeq \Mm^{-1}$ (i.e., that matrix $\Sigmam - \Mm^{-1}$ is positive
semidefinite), $\Mm$ being the Bayesian information matrix (BIM)
defined by
\begin{equation}
  \label{eq:info_matrix} \Mm \triangleq -\EE_{\Wc, \thetav} \left[
  \nabla_{\thetav} \nabla_{\thetav} \Tran \log \PP\left( \Wc,
    \thetav\right) \right]
\end{equation}
where $\nabla_{\thetav}$ represents the gradient with respect to the vector of parameters $\thetav$. 
The BCRB implies that:
\begin{equation} [\Sigmam]_{i,i} \geq [\Mm^{-1}]_{i,i}
\end{equation}
i.e., it provides a lower bound to the MSE achieved by any estimator of the $i$-th unknown parameter, $i=1,\ldots,N+K$.  The following proposition gives the BCRB for our
scenario.

\begin{prop}\label{prop:BCRB}
For the above described scenario the BIM is given by
\beq
\Mm = \left[ 
\begin{array}{cc}
\Gammam \Deltam_q \Gammam\Tran + \beta_q \Id_N & \zerov_{N \times K} \\
\zerov_{K \times N} & \Deltam_{\rho} + \beta_{\rho} \Id_K
\end{array}
\right]\label{eq:BIM}
\eeq
where $\Gammam$ is defined through \eqref{eq:d_versus_q}, $\Deltam_q$ and $\Deltam_{\rho}$ are diagonal matrices with diagonal elements
\beq \label{eq:Delta_o}
[\Deltam_q]_{e,e} = |\Kc_e| \,\EE_{\rho,d} \left[ \frac{ \rho^2(F'(\rho d))^2}{F(\rho d) (1-F(\rho d))} \right], e=1,\dots,E
\eeq
and
\beq \label{eq:Delta_w}
[\Deltam_{\rho}]_{k,k} = |\Ec_k | \,\EE_{\rho,d} \left[ \frac{ d^2(F'(\rho d))^2}{F(\rho d) (1-F(\rho d))} \right], k = 1,\dots,K
\eeq
respectively. Finally:
\beq
 \beta_q{=} -\EE_q\left[\frac{\partial^2 }{\partial q^2} \log f_q(q)\right]\mbox{ and } \beta_{\rho}{=} -\EE_\rho\left[\frac{\partial^2 }{\partial \rho^2} \log f_{\rho}(\rho)\right].
\eeq
All averages are performed with respect to marginals of qualities, quality distances or reliabilities.
\end{prop}

\begin{IEEEproof}
See Appendix~\ref{app:proof_BCRB}.
\end{IEEEproof}
Notice that, since the BIM is block diagonal, the BCRB on qualities
and the BCRB on reliabilities can be computed separately, by inverting
the two diagonal blocks. Moreover, the lower block is diagonal, so the
BCRB on worker reliabilities can be explicitly written as
\begin{equation}
  \label{eq:bcrb_rel} \EE \left[|\widehat{\rho}_k -\rho_k |^2\right] \geq  ([\Deltam_{\rho}]_{k,k} + \beta_{\rho})^{-1}
\end{equation}

Regarding the MSE on quality distances, it can be found that 
\begin{eqnarray}
\EE \left[(\widehat{\dv} - \dv ) (\widehat{\dv} - \dv )\Tran \right] & = & \Gammam\Tran \EE \left[(\widehat{\qv} - \qv ) (\widehat{\qv} - \qv )\Tran \right] \Gammam \non
& \succeq & \Gammam\Tran \left( \Gammam \Deltam_q \Gammam\Tran + \beta_q \Id_N \right)^{-1} \Gammam \nonumber
\end{eqnarray}

Finally, we particularize the above computations for the two worker
models considered in this paper. For the Thurstone model, the argument
of the average in~\eqref{eq:Delta_o} and~\eqref{eq:Delta_w} can be
rewritten as
\beq \frac{(F'(\rho d))^2}{F(\rho d) (1-F(\rho d))} = \frac{2}{\pi} \frac{\ee^{-\rho^2 d^2}}{1 - \erf^2 \left( \rho d     /\sqrt{2}\right)} \eeq
while for the BTL model we have
\beq \frac{(F'(\rho d))^2}{F(\rho d) (1-F(\rho d))} = \frac{\ee^{\rho d}}{(1 + \ee^{\rho d})^2} \eeq

\section{Ranking algorithms for heterogeneous workers\label{sec:ranking_algorithms}}
In this section, first we briefly review the most relevant classes of
algorithms proposed in literature: the LS algorithms proposed
in~\cite{noiTNSE} for an homogeneous case, which constitute the kernel
around which our proposed QUITE algorithm has developed, and the AG algorithm proposed
in~\cite{cinesi}, which is, to the best of our knowledge, the only
proposed ranking algorithm explicitly tailored for the heterogeneous case.
Then,  we introduce the single-stage version of QUITE algorithm.
\subsection{Algorithms from the literature}\label{sect:previous_algo}

First, we briefly recall a family of algorithms based on Least-Square
(LS) estimation of object qualities, which have been introduced
in~\cite{noiTNSE} for the homogeneous case, i.e., when all workers are
characterized by the same reliability $\rho$. These algorithms start
from an initial unbiased estimate of distance $d_e$, given by
$\widehat{d}_e=\frac{1}{\rho}F^{-1}(\widehat{p}_e)$, where
$\widehat{p}_e=1-|\Kc_e|^{-1}\sum_{ k\in \Kc_e}w_{e,k}$ is an unbiased
estimate of the probability $F(\rho d_e)$.  From the noisy estimates
$\hat{\dv} =[\hat{d}_1,\ldots, \hat{d}_E]\Tran$, the estimates
$\widehat{\qv}=[\widehat{q}_1,\ldots,\widehat{q}_N]\Tran$ of
$\qv=[q_1,\ldots,q_N]\Tran$ are then obtained by solving the following
LS optimization problem
\begin{equation}\label{min-prob}
\widehat{\qv} = \arg \min_{\xv}  \sum_{e \in \Ec} \omega_e\left(x_{i_e}-x_{j_e} - \widehat{d}_{e} \right)^2 
\end{equation}
where $ \omega_e$ are arbitrary positive weights,  whose setting  is discussed in \cite[Sect. 3.1]{noiTNSE}.

An algorithm that specifically targets ranking with heterogeneous
workers is described in~\cite{cinesi}. It implements  an alternate-gradient (AG) optimization, which approximates joint ML estimation of
objects qualities and worker reliabilities. More precisely, by
exploiting \eqref{eq:d_versus_q}, let
\beq \Lc(\rhov, \qv) = -\log \PP
(\Wc | \rhov, \Gammam\Tran \qv)\,. \eeq
The algorithm in~\cite{cinesi} starts from an arbitrary point in the $(\rhov, \qv)$ space, call it
$(\widehat{\rhov}^{(0)}, \widehat{\qv}^{(0)})$ and works iteratively. At iteration
$\ell$, $\ell = 1,2,\dots$ it performs the following three steps:
\begin{eqnarray}
 \widetilde{\qv}^{(\ell)} & = & \widehat{\qv}^{(\ell-1 )} - \lambda_q \nabla_{\qv} \Lc(\rhov, \qv) \Big|_{\widehat{\rhov}^{(\ell-1)}, \widehat{\qv}^{(\ell-1)}} \label{eq:q_step}\\
 \widehat{\qv}^{(\ell)} & = & \widetilde{\qv}^{(\ell)} - \frac1{N} \mathbf{1}_{N}\Tran \widetilde{\qv}^{(\ell)} \\
 \widehat{\rhov}^{(\ell)} & = & \widehat{\rhov}^{(\ell-1 )} - \lambda_{\rho} \nabla_{\rhov} \Lc(\rhov, \qv) \Big|_{\widehat{\rhov}^{(\ell-1)}, \widehat{\qv}^{(\ell-1)}} \label{eq:rho_step}
\end{eqnarray}
where $\lambda_q$ and $\lambda_{\rho}$ are positive step sizes. Iterations stop when a predetermined maximum number of iterations have been performed.

%
%
\begin{figure}[t]
\centerline{\includegraphics[width=1.00\columnwidth]{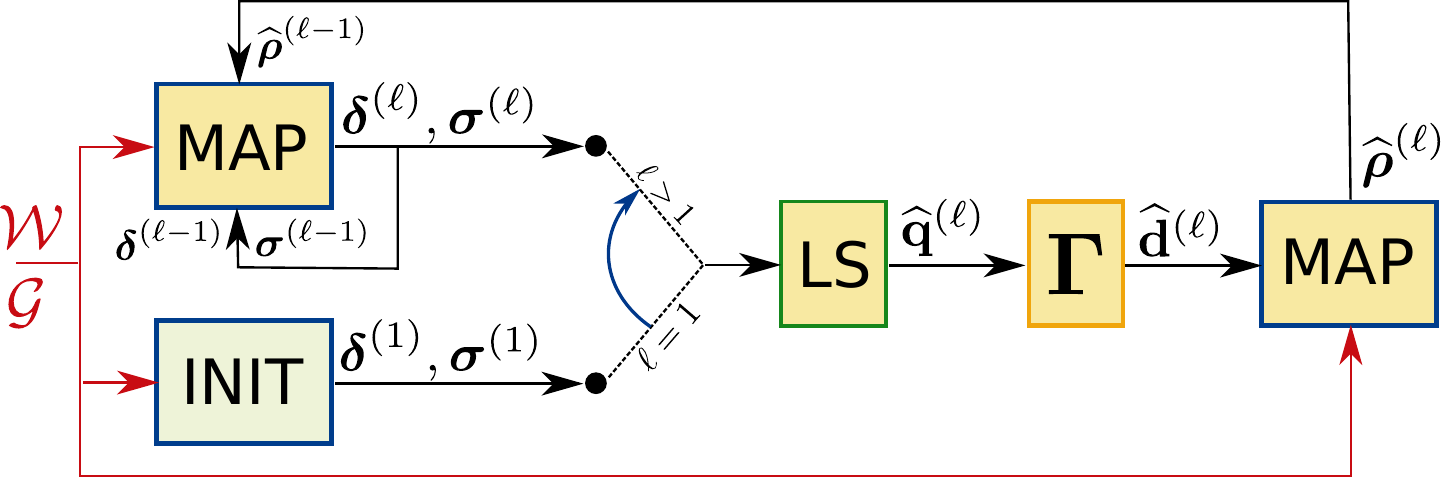}}
\caption{Block scheme of the QUITE algorithm. The variable
  $\ell$ denotes the iteration index.  The algorithm takes as input
  the set of workers' answers $\Wc$ and the graph $\Gc$ and, at each iteration, outputs
  the quality estimates $\widehat{\qv}^{(\ell)}$, and the estimates of
  workers reliabilities $\widehat{\rhov}^{(\ell)}$. Iterations stop when
  condition~\eqref{eq:QUITE_stop} is met or when the maximum number of
  iterations, $I_{\rm max}$, is reached.  }\label{fig:figure_QUITE}
\end{figure}
\subsection{The QUITE algorithm\label{sec:estimation}}

In this subsection, we propose an algorithm named QUality ITerative
Estimator (QUITE), which iteratively refines the estimation of quality
distances and worker reliabilities. It consists in an alternate
optimization of $\dv$ given $\rhov$ and of $\rhov$ given $\dv$, along
a total number of $I_{\max}$ iterations, with the possibility of early
termination thanks to a stopping condition.  In the following, the
symbol $\ell$ is used to denote the iteration index and the value of
the generic variable $v$ at iteration $\ell$ is indicated as
$v^{(\ell)}$.

Algorithm~\ref{alg:QUITE-strong} reports the pseudo-code for QUITE whereas Figure~\ref{fig:figure_QUITE} shows its simplified block scheme. The algorithm
takes as inputs the set of workers' answers $\Wc$, the graph $\Gc$, the workers' model
$F(\cdot)$, the graph incidence matrix $\Gammam$, the a-priori
distributions on object qualities, distances and workers'
reliabilities $f_q(\cdot)$, $f_{d}(\cdot)$ and $f_{\rho}(\cdot)$,
respectively, and their support $\Ic_q$, $\Ic_d$ and
$\Ic_{\rho}$. QUITE also takes as input the maximum number of
iterations $I_{\max}$ and a threshold $\tau$ for the stopping
condition.

The QUITE algorithm initializes the estimates of the object qualities,
$\widehat{\qv}^{(0)}$, by randomly drawing it from the distribution
$f_q(q)$, $q\in \Ic_q$.  Then,  at iteration $\ell = 1,2,\dots $,  the following operations are performed.
\begin{itemize}
\item First QUITE infers rough estimates of the edge distances,
  $\deltav^{(\ell)}=[\delta_1^{(\ell)},\ldots,\delta_E^{(\ell)}]\Tran$,
  and of their variances
  $\sigmav^{(\ell)}=[\sigma_1^{(\ell)},\ldots,\sigma_E^{(\ell)}]\Tran$,
  on a per-edge basis. Such procedure is detailed in
  Section~\ref{sec:est_d}; it exploits the workers' answers, $\Wc$,
  and the estimates of the workers' reliabilities at previous
  iteration, $\widehat{\rhov}^{(\ell-1)}$, when available.

\item Then, the estimates $\deltav^{(\ell)}$ and their variances, $\sigmav^{(\ell)}$,
are combined together through a weighted LS algorithm exploiting the graph, in
order to obtain estimates $\widehat{\qv}^{(\ell)}$ of the object
qualities, as described in Section~\ref{sec:graph_LS}.
\item Next, new estimates of the distances, $\widehat{\dv}^{(\ell)}$,
are finally obtained from $\widehat{\qv}^{(\ell)}$ through the incidence
matrix $\Gammam$, by applying~\eqref{eq:d_versus_q}.  

\item Finally, MAP estimates of the workers' reliabilities,
  $\widehat{\rhov}^{(\ell)}$, are obtained from
  $\widehat{\dv}^{(\ell)}$, as detailed in Section~\ref{sec:est_rho}.
\end{itemize}

The algorithm stops when the normalized difference
$\widehat{\qv}^{(\ell)}-\widehat{\qv}^{(\ell-1)}$ is sufficiently
small, i.e., when
\begin{equation}
  \frac{\left\|\widehat{\qv}^{(\ell)}-\widehat{\qv}^{(\ell-1)}\right\|_2}{N \left\|\widehat{\qv}^{(\ell-1)}\right\|_2} < \tau
  \label{eq:QUITE_stop}
\end{equation}
or when the maximum number of iterations $I_{\rm max}$  is reached.

\subsubsection{Estimates of the edge distances and of their variance\label{sec:est_d}}
Here, we describe how the estimates $\deltav^{(\ell)}$ of the distances
$\dv$ are obtained by the QUITE algorithm.  At
the first iteration ($\ell=1$),  since estimates of the workers'
reliabilities are not yet available, we resort to a
generalization of the procedure employed in~\cite{noiTNSE} and briefly recalled in Sect. \ref{sect:previous_algo}, taking
into account that workers have different reliabilities.  Specifically,  for edge $e$, 
we first estimate the empirical probability of a 0 answer as
\begin{equation}
  \widehat{p}_e=1-|\Kc_e|^{-1}\sum_{k\in \Kc_e}w_{e,k}\,.
  \label{eq:initial_pe}
\end{equation}
If all workers have the same reliability, $\rho$,
using~\eqref{eq:P_w0}, the estimate of the edge distance $\delta_e$
can be obtained by
$\delta_e^{(1)}=\frac{1}{\rho}F^{-1}(\widehat{p}_e)$, $e=1,\ldots,
E$.  When workers have different random reliabilities, whose priors
are $f_{\rho}(\rho)$, the estimate of the edge distance can be
generalized as
\begin{equation}
  \delta_e^{(1)}= G^{-1}(\widehat{p}_e)
  \label{eq:delta_e_initial}
\end{equation}
where the function $G(\delta)$ is given by
\begin{equation}
  G(\delta)= \int_{\Ic_{\rho}}F(\rho \delta) f_{\rho}(\rho) \dd \rho
  \label{eq:G}
\end{equation}
The variance of such estimate is computed as~\cite[Section 3.1]{noiTNSE}
\beq \label{eq:sigma_e_initial}
\sigma^{(1)}_{e} = \left(\left.\frac{d G^{-1}(p)} {d p}\right|_{p=\widehat{p}_e} \right)^2  
\frac{\widehat{p}_e(1-\widehat{p}_e)}{|\Kc_e|}\,.
\eeq
The above procedure is denoted by the block labeled ``INIT'' in Figure~\ref{fig:figure_QUITE}.

For $\ell>1$, since estimates $\widehat{\rhov}^{(\ell-1)}$ of the workers'
reliabilities are available we can apply a ``MAP'' approach to
obtain the estimates $\deltav^{(\ell)}$.
%
%
Let us focus on object pair $e$ and let $\Wc_e$ be the
set of workers' answers related to such pair.  Then, according
to~\eqref{eq:pw}, the log a-posteriori probability of the distance
$d_e$, given the workers' reliabilities $\rhov$ and answers $\Wc_e$ is
given by
  \begin{eqnarray}
    &&\hspace{-10ex}\log \PP(d_e| \Wc_e, \rhov) \non
    &=& \log\frac{\PP(\Wc_e | d_e, \rhov)f_d(d_e)}{\PP(\Wc_e)} \non
    &=& \sum_{k\in \Kc_e}\log F(x_{e,k}){+}\log f_d(d_e){-}\log \PP(\Wc_e)\label{eq:a_posteriori}
  \end{eqnarray}
  where  we recall that $x_{e,k}\triangleq (1-2w_{e,k})\rho_kd_e$.
  At iteration $\ell$, since the true reliabilities $\rhov$ are
  unknown and only their estimates $\widehat{\rhov}^{(\ell-1)}$ are available, according
  to~\eqref{eq:a_posteriori} we can write
  \begin{equation}
    \delta_e^{(\ell)}{=} \arg\max_{d\in \Ic_d}\sum_{k\in \Kc_e}\log F\left((1{-}2w_{e,k})\widehat{\rho}_k^{(\ell-1)} d \right) {+}\log f_{d,e}^{(\ell)}(d)\,.
    \label{eq:delta_e2}
  \end{equation}
 Moreover, since at step $\ell-1$ the random
variable $d_e$ has been estimated having mean $\delta_e^{(\ell-1)}$ and
variance $\sigma_e^{(\ell-1)}$, the prior for $d_e$ 
has been replaced in~\eqref{eq:delta_e2} with a Gaussian distribution
with such mean and variance, i.e., $f_{d,e}^{(\ell)}(d) = \Nc(d; \delta_e^{(\ell-1)},
\sigma_e^{(\ell-1)})$. 
The maximization involved in~\eqref{eq:delta_e2} must be  performed numerically
since, for both Thurstone and BTL models, it is not
solvable analytically. However, for both models, the function $F(\cdot)$ is
log-concave, so that there is a single minimum, which can be found by
efficient numerical methods.

The variance of the estimate can be obtained starting
 from~\eqref{eq:a_posteriori}. Note that, in order to
 maximize~\eqref{eq:a_posteriori}, we can set to zero its derivative
 w.r.t. to $d_e$, i.e., we compute
\begin{eqnarray}
  &&\hspace{-5ex}R(d_e)\non
  &=& \frac{\partial }{\partial d_e} \left(\sum_{k\in \Kc_e} \log F(x_{e,k}) {+} \log f_d (d_e) {-}\log \PP(\Wc_e)\right)\non
  &=& \sum_{k \in \Kc_e} (1-2w_{e,k})\rho_k \frac{F'(x_{e,k})}{F(x_{e,k})}  + \frac{f_d'(d_e)}{f_d(d_e)} = 0\label{eq:derivative_a_posteriori}
\end{eqnarray}
Let $d_e^*$ be the solution of~\eqref{eq:derivative_a_posteriori}. Then, $d_e^*$ can be seen as an implicit function of the arguments $w_{e,k}$, $e\in \Kc_e$.
With an abuse of notation, by considering $w_{e,k}$ as a continuous variable, we can approximate $d_e^*$ as $d_e^*\approx \sum_{k \in \Kc_e} \frac{\partial d_e^*}{\partial w_{e,k}} w_{e,k}$. Since the random variables $w_{e,k}$ are independent we can write
\begin{equation} \label{eq:first_order_var}
\VV[d_e^*] \approx \sum_{k \in \Kc_e} \left(\frac{\partial d_e^*}{\partial w_{e,k}} \right)^2 \VV[w_{e,k}]\,,
\end{equation}
where we recall that $\VV[\cdot]$ is the variance operator and $w_{e,k}$ is a Bernoulli random variable with parameter $F(\rho_kd_e^*)$. Therefore
\begin{equation} \label{eq:var_w}
 \VV[w_{e,k}] = F(\rho_k d_e^*) \left[1-F(\rho_k d_e^*)\right]\, .
\end{equation}
Moreover,  thanks to implicit differentiation, we can obtain the ``derivative'' in \eqref{eq:first_order_var} as
\begin{equation} \label{eq:frac_der}
\frac{\partial d_e^*}{\partial w_{e,k}} = -\frac{\Delta R(d_e^*) / \Delta w_{e,k}}{\partial R(d_e^*) / \partial d_e^*}.
\end{equation}
The numerator of~\eqref{eq:frac_der} can be expressed through the difference quotient, given by
\begin{eqnarray}
  \frac{\Delta R(d_e^*)}{\Delta w_{e,k}}
  &=& R(d_e^*)\big|_{w_{e,k} = 1} - R(d_e^*)\big|_{w_{e,k} = 0}\non
  &=& -\frac{\rho_k F'(\rho_k d_e^*)}{F(\rho_k d_e^*) (1-F(\rho_k d_e^*))} \label{eq:frac_num}
\end{eqnarray}
The result in~\eqref{eq:frac_num} has been obtained by using~\eqref{eq:derivative_a_posteriori} and by observing that, for models as
in~\eqref{eq:F_Thurstone} and~\eqref{eq:F_BTL}, we have $F(-x)=1-F(x)$ and $F'(x) = F'(-x)$. The denominator of~\eqref{eq:frac_der} is trivially given by 
\begin{eqnarray}
  \frac{\partial R(d_e^*)}{\partial d_e^*} &=& \sum_{k \in \Kc_e} \rho_k^2 \frac{F''(x^*_{e,k}) F(x^*_{e,k}) - F'(x^*_{e,k})^2}{F(x^*_{e,k})^2}\non
  &&\qquad\qquad+ \frac{f_d''(d_e^*) f_d(d_e^*) - f_d'(d_e^*)^2}{f_d(d_e^*)^2} \label{eq:frac_den} \\
  &\triangleq& u(d_e^*)
\end{eqnarray}
where we have defined for brevity
$x_{e,k}^*\triangleq (1-2w_{e,k})\rho_k
d_e^*$. Substituting \eqref{eq:var_w}-\eqref{eq:frac_den} into
\eqref{eq:first_order_var}, we obtain 
\begin{equation}
  \VV[d_e^*] \approx\frac{\displaystyle\sum_{k \in \Kc_e} \rho_k^2 \frac{ F'(\rho_k d_e^*)^2}{F(\rho_k d_e^*) (1-F(\rho_k d_e^*))}}{\left(u(d_e^*)\right)^2}
  \end{equation}
The estimated variance for object pair $e$ at iteration $\ell$ is then given by 
  \begin{equation}
  \sigma_e^{(\ell)} = \VV[d_e^*]\big|_{d_e^* = \delta_e^{(\ell)}, \,\,\rho_k = \widehat{\rho}_k^{(\ell-1)}}
  \label{eq:sigma_e2}
  \end{equation}

\begin{algorithm}
\caption{QUITE algorithm} \label{alg:QUITE-strong}
\begin{algorithmic}[1]
  \REQUIRE{$\Wc$, $\Gammam$, $F(\cdot)$, $f_q(\cdot)$, $f_d(\cdot)$, $f_{\rho}(\cdot)$, $\Ic_q$, $\Ic_d$, $\Ic_{\rho}$, $I_{\max}$, $\tau$}
  \ENSURE{$\widehat{\qv}$}
  
  \STATE initialize $\widehat{\qv}^{(0)}$ with i.i.d. random entries according to the distribution $f_{q}(q)$, $q\in \Ic_q$
  \FOR{$\ell = 1, 2, \dots, I_{\max}$}
    \IF{$\ell=1$}
       \FOR{$e = 1, 2, \dots, E$}
  \STATE Compute $\widehat{p}_e$ and $\delta_e^{(1)}$ according to~\eqref{eq:initial_pe} and~\eqref{eq:delta_e_initial},   respectively. \label{step:init}     
       \STATE Compute $\sigma_e^{(1)}$ according to~\eqref{eq:sigma_e_initial}
       \ENDFOR
  \ELSE
  \FOR{$e = 1, 2, \dots, E$}
  \STATE  Compute $\delta_e^{(\ell)}$ using~\eqref{eq:delta_e2} and $\sigma_e^{(\ell)}$ using~\eqref{eq:sigma_e2} \label{eq:d_est}
    \ENDFOR
  \ENDIF
\STATE Use the weighted LS algorithm of~\cite{noiTNSE}, and compute $\widehat{\qv}^{(\ell)} = \mathrm{LS} \left(\deltav^{(\ell)}, \sigmav^{(\ell)}\right)$.
\STATE Update the estimate of $\dv$ using~\eqref{eq:d_versus_q}: $\widehat{\dv}^{(\ell)}= \Gammam\Tran \widehat{\qv}^{(\ell)}$

\FOR{$k=1,\dots,K$}
    \STATE Compute $\widehat{\rho}_k^{(\ell+1)}$ using~\eqref{eq:rho_k2}  
  \ENDFOR

\IF{$\left\|\widehat{\qv}^{(\ell)} - \widehat{\qv}^{(\ell-1)}\right\|_2< \tau N \left\|\widehat{\qv}^{(\ell-1)}\right\|_2$} \label{step:stop}
  \STATE {\bf break}
\ENDIF

\ENDFOR 
\STATE \textbf{return} $\widehat{\qv} = \widehat{\qv}^{(\ell)}$

\end{algorithmic}
\end{algorithm}

\subsubsection{Graph estimation of quality distances\label{sec:graph_LS}}
We now describe the LS algorithm that is used in
Alg.~\ref{alg:QUITE-strong}, to obtain the quality estimates
$\widehat{\qv}^{(\ell)}$, given current estimates of quality distances
$\deltav^{(\ell)}$ and of their variances $\sigmav^{(\ell)}$.  In
particular, we extend the approach introduced in~\cite{noiTNSE}, where
the quality estimates are obtained by solving the weighted LS problem
\begin{equation}\label{eq:LS}
  \widehat{\qv}^{(\ell)}  = \arg\min_{\xv}\sum_{e \in \Ec} \omega_e^{(\ell)} \left( x_{i_e}-x_{j_e}- \delta_e^{(\ell)}\right)^2 = {\rm LS}(\deltav^{(\ell)},\sigmav^{(\ell)})
\end{equation}
where $\omega_e^{(\ell)}$ are arbitrary positive edge weights. The problem can
be solved by exploiting the graph structure as described in~\cite[Equation (13)]{noiTNSE}.  
With the assumption that per-edge distance estimates are independent, the weights are chosen in order to minimize the variance of the quality
  estimates,  i.e.
  \begin{equation} \label{eq:var_est_dist}
    \omega_e^{(\ell)} = \frac1{ \sigma_e^{(\ell)}}.
  \end{equation}

\subsubsection{Estimates of the workers' reliabilities\label{sec:est_rho}}
The procedure for estimating the reliabilities $\rhov$ is similar to
that employed to estimate the edge distances.
Again, we use a MAP approach, i.e., for worker $k$, we maximize the log a-posteriori probability 
 \begin{eqnarray}
   && \hspace{-10ex}\log \PP(\rho_k| \Wc_k, \dv) \non
    &=& \log\frac{\PP(\Wc_k | \rho_k, \dv)f_\rho(\rho_k)}{\PP(\Wc_k)} \non
   &=& \sum_{e\in \Ec_k}\log F(x_{e,k}){+}\log f_\rho(\rho_k){-}\log \PP(\Wc_k)\label{eq:a_posteriori_rho}
 \end{eqnarray}
where we recall that $\Ec_k$ is the set of object pairs evaluated by
worker $k$ and $\Wc_k$ are the corresponding evaluation answers.
  Since the true distance $\dv$ are unknown and only their estimates $\dv^{(\ell)}$ are available,
  according to~\eqref{eq:a_posteriori},  we can write
  \begin{equation}
    \rho_k^{(\ell)} = \arg\max_{\rho\in \Ic_\rho}\sum_{e\in \Ec_k}\log F\left((1-2w_{e,k})d_e^{(\ell)} \rho \right) +\log f_{\rho}(\rho)\,.
    \label{eq:rho_k2}
  \end{equation}
\subsection{Theoretical guarantees for QUITE}
 
 The  asymptotical optimality of the LS algorithm is preserved also in the heterogeneous
 case.  In particular we can claim that: 
 
 \begin{proposition} \label{prop:theor}
 A single-iteration version of the QUITE algorithm,  according to which  initial distances estimates $\delta_e$ are computed as in \eqref{eq:delta_e_initial},  and then 
 	 quality estimates are obtained  by solving the optimization problem   \eqref{min-prob},  with weights 
 	 $ \omega_{i,j}=1$, is asymptotical optimal as long as $\inf_x \frac{\mathrm d G(x)}{\mathrm d
 	 	x}>0$.
 \end{proposition}
In Appendix \ref{extension-proofs} we report a brief
 discussion of how proofs in \cite{noiTNSE} can be extended to the heterogeneous case.
\section{Two-stage QUITE algorithm\label{sec:2stage}}

\begin{figure}[t]
\centerline{\includegraphics[width=1.00\columnwidth]{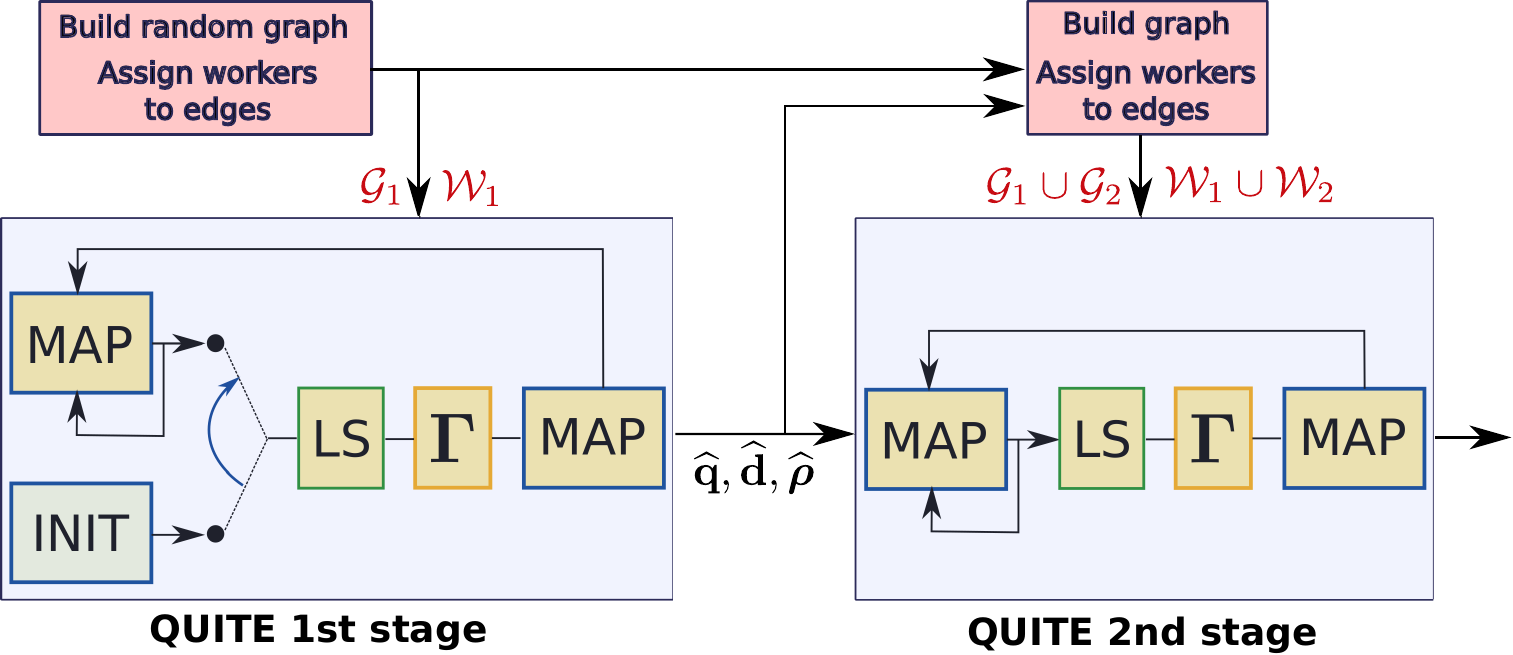}}
\caption{Block scheme of the two-stage QUITE algorithm. The first stage
  is as in Figure~\ref{fig:figure_QUITE} whereas the second stage has
  been slightly simplified by removing the block ``INIT''. The output
  of the first stage is used to build a new graph $\Gc_2$, and to
  initialize the second stage.}\label{fig:figure_QUITE2}
\end{figure}

In this section, we describe a two-stage protocol employing our QUITE
algorithm as the fundamental and whose goal is to
improve the reliability of the estimated ranking. Its block scheme is depicted in Figure~\ref{fig:figure_QUITE2}. The first stage
works exactly as previously described and works on a graph
$\Gc_1$ and on the set of answers $\Wc_1$. After obtaining an estimate of object qualities, $\widehat{\qv}$, and
worker reliabilities, $\widehat{\sigmav}$, a second stage consists in the following three
steps:
\begin{enumerate}
\item a new graph $\Gc_2$ is created, with objects as nodes and edges
  in one-to-one correspondence to object pairs to be evaluated;

\item object pairs thus obtained are sent out for evaluation to the
  same set of workers that have evaluated the object pairs in the
  first stage;

\item after collecting the evaluations $\Wc_2$, the QUITE algorithm is
  executed again on the joint set of evaluations collected in the two
  stages, i.e., $\Wc_1 \cup \Wc_2$ to yield an improved ranking.
\end{enumerate}  

In the second stage, the QUITE algorithm has only a slight difference
with respect to the version described in Section \ref{sec:estimation}.
Indeed, instead of initializing the distance estimates as in line
\ref{step:init} of Algorithm \ref{alg:QUITE-strong}, the algorithm
simply takes the estimates obtained at the output of the first stage and use
them to find an estimate of distances, as in line \ref{eq:d_est} of
the same algorithm. Then such distance estimates are used to obtain an
improved estimate of worker reliabilities, and so on.  In practice, in
the second stage the block labeled ``INIT'' is removed, since it is
unnecessary.

What makes the second stage a breakthrough is the fact that both the graph
$\Gc_2$ and the assignment of object pairs to workers build upon
the results of the first stage. In the two following subsections, we
explain in detail these procedures.

\subsection{Building the second stage graph\label{sec:build_G2}} 

While graph $\Gc_1$ in the first stage must be chosen as a random
(regular) graph, the knowledge obtained in the first stage can be used
to drive the choice of $\Gc_2$. The idea is that the most
important information for a reliable ranking estimation comes from
evaluating pairs of objects that are close in the true
ranking. Inspired by such idea, we propose to form graph $\Gc_2$
as follows.

Let $\widehat{\qv}$ be the estimates obtained at the output of the first
stage and let $\widehat{\piv}$ be the permutation of integers
$\{1,\dots,N\}$ such that $\widehat{q}_{\widehat{\pi}_1} >
\widehat{q}_{\widehat{\pi}_2}> \dots > \widehat{q}_{\widehat{\pi}_N}$,
i.e., $\widehat{\piv}$ is essentially the ranking estimated at the
first stage.  Then, given a maximum even node degree $D$, to build the graph $\Gc_2$ we proceed as follows:
for $i=1,\ldots,N$ we define the set
\begin{equation}
  \Sc_i=\left\{ i'\in\{1\ldots,N\}\setminus i \Big| |i-i'| \leq \frac{D}{2}\right\}
\end{equation}
and in $\Gc_2$ we connect with an edge objects $\widehat{\pi}_i$ with
all objects $\widehat{\pi}_{i'}$, $i'\in\Sc_i$. 
By doing so, node $\widehat{\pi}_i$
is connected (in $\Gc_2$) with at most $D$ of its closest neighbors in the estimated ranking $\widehat{\qv}$.   
Note that the cardinality of $\Sc_i$ depends on $i$. Indeed, objects
located at the top or at the bottom of the estimated ranking $\widehat{\piv}$ are
expected to have a smaller number of neighbors with respect to objects
in the middle of the ranking list. Therefore, the graph $\Gc_2$ is not regular.

\subsection{Assignment of object pairs to workers\label{sec:recipe}}
Once the graph $\Gc_2$ is built, we need to assign object pairs to the
pool of workers. For such assignment we must exploit the information
we gathered in the first stage about the workers' reliabilities, i.e., $\widehat{\rhov}$. 

Consider a particular object pair, $e$, in the graph $\Gc_2$ and
assume that it is assigned to a set of workers $\Kc_e$, all having the
same reliability $\rho$.  In this simplified case, we recall that the
estimate of the distance $d_e$ based on the workers' answers,
$w_{e,k}$, $k \in \Kc_e$, is given by
$\widehat{d}_e=\frac{1}{\rho}F^{-1}(\widehat {p}_e)$, where
$\widehat{p}_e = 1-\frac{1}{|\Kc_e|}\sum_{k\in \Kc_e} w_{e,k}$ is the
unbiased estimate of the probability $p_e$, i.e., $\EE[\widehat{p}_e]
=p_e$.
Therefore, the estimation error $y_e =\widehat{p}_e-p_e$ is a
zero-mean random variable with variance
\begin{equation}
  \VV[y_e]
  = \EE[(\widehat{p}_e-p_e)^2]  = \EE[\widehat{p}_e^2]-p_e^2   = \frac{p_e(1-p_e)}{|\Kc_e|}
\end{equation}
Correspondingly, let $z_e=\widehat{d}_e -d_e$ the estimation error on the distance $d_e$. 
For small $y_e$, $z_e$
is a random variable with zero mean and variance
\begin{eqnarray} \VV[z_e]
  &\approx& \left(\frac1{\rho}\frac{\dd F^{-1}}{\dd x}\Big|_{x= p_e} \right)^2\VV[y_e]\non
  &=&\left( \frac1{\rho}\frac{\dd F^{-1}}{\dd x}\Big|_{x= p_e}\right)^2\frac{p_e(1-p_e) } { |\Kc_e|}
\label{eq:var_ze}
\end{eqnarray}
Now, under the BTL model, by using~\eqref{eq:F_BTL} in~\eqref{eq:var_ze} we have:
\[
\VV[z_e]\approx H(\rho)= \frac{1}{\rho^2} \frac{[ \ee^{-\rho d_e}+2+ \ee^{\rho d_e}]}{ |\Kc_e|}= \frac{2}{\rho^2} \frac{ \cosh(\rho d_e)+1}{ |\Kc_e|}
\]
Note that $H(0)=H(\infty)=\infty$, and therefore the optimal value
$\rho^*_e$ of $\rho$ for pair $e$, which is the one that minimizes
$\VV[z_e]$, is not extremal and can be obtained by imposing
$\frac{\dd H(\rho) } {\dd \rho}= 0$, yielding
$\rho^*_e \approx \frac{2.399}{|d_e|}$. A similar derivation can be
carried out as well for the Thurstone model. Then, denoting
$\widehat{\rho}^*_e = \rho^*_e \big|_{d_e = \widehat{d}_e}$, (where $\widehat{d}_e$ is an estimate of $d_e$ provided by the first QUITE stage) the rule
we propose for the assignment of object pairs to workers is as
follows: assign workers to edges so to minimize
the metric $\sum_{k=1}^K \sum_{e \in \Ec_k}|\widehat{\rho}_k-
\widehat{\rho}^*_e|$.  Such assignment can be easily implemented by
complying with the following simple rule: ``assign best workers to
shortest links''.

\section{Numerical results\label{sec:results}}

In this section, we first provide numerical results of the performance
of the single-stage QUITE algorithm and compare it with the AG
algorithm proposed in~\cite{cinesi}.  Subsequently, we show the
performance improvement achieved by the two-stage QUITE algorithm.

\subsection{Single-stage QUITE}
To show the performance of QUITE we rely on two metrics: the achieved
MSE on the estimated object qualities and the error probability of the
estimated ranking.

In our simulations we consider a number of objects, $N$, ranging from
40 to 400 and a set of workers of size $K=N$. For each value of $N$,
we build a random regular graph with degree $D = 20$, which is kept
fixed for the whole experiment. The graph has then $E= N D/2 = 10N$
edges. Each of these corresponds to an object pair evaluated by $M =
\alpha K$ ($\alpha\le 1$) different workers, with $\alpha \in \{ 0.1,
0.2, 0.5, 1\}$. Allocation of workers to object pairs is regular, so
that each worker evaluates exactly $EM/K=\alpha E$ object pairs.

Object qualities and worker reliabilities are i.i.d. and drawn from
the uniform distributions $f_q(q) = \Uc[0,1]$ and $f_\rho(\rho) =
\Uc[1,20]$, respectively, which are supposed to be known by the
algorithm.

Since such distribution functions have discontinuities, for the
computation of the BCRB we have windowed it with the
Planck-taper~\cite{Planck-Taper} function which approximates the
uniform distribution $\Uc[a,b]$ with a finite-support probability
density function of class $C^{\infty}$, given by
$$
f_{z}^{\mathrm{Planck}}(x) = \left\{
\begin{array}{ll}
\frac{C_p}{1 + \exp\left\{z \left( \frac1{x-a} - \frac1{a+z-x}\right) \right\}}, & a{<}x{<}a{+}z, \\
C_p, &  a{+}z {\leq} x {\leq} b{-}z, \\
\frac{C_p}{1 + \exp\left\{-z \left( \frac1{b-z-x} - \frac1{x-b}\right) \right\}}, & b{-}z{<}x{<}b, \\
0, & \mathrm{elsewhere}
\end{array}
\right.$$
where $C_p{=}1/(b{-}a{-}z)$, the parameter $0{<}z{\le}(b{-}a)/2$ represents the
smoothness of the function.  In the BCRB computation, we have chosen
$z{=}(b{-}a)/5$ to reasonably approximate the true a-priori
distributions.

\subsubsection{Mean square error on the quality estimates} 
\begin{figure}[t]
\centerline{\includegraphics[width=1.00\columnwidth]{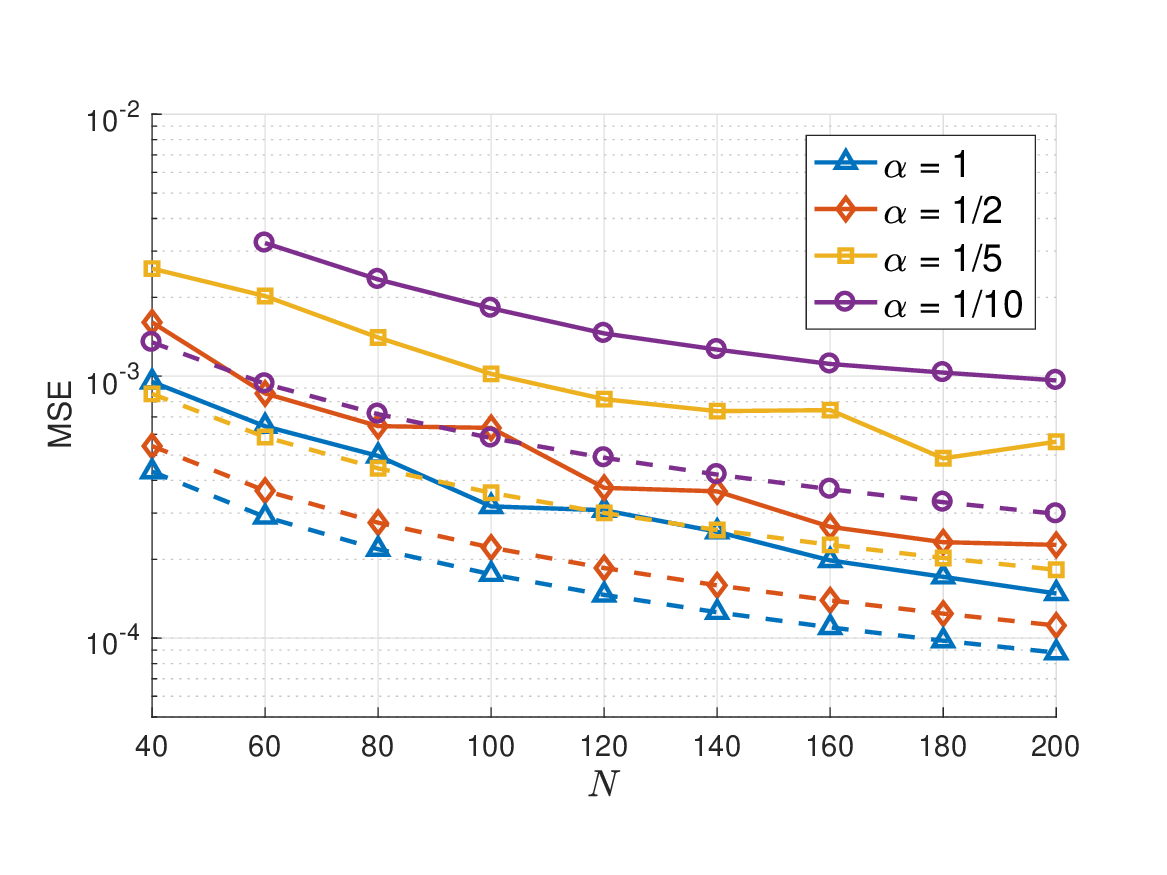}}
\caption{MSE performance of the QUITE algorithm versus the number of
  objects for $K = N$ workers, and BTL worker model. The graph is
  regular and has degree $D = 20$. $M = \alpha K$ evaluations
  are carried out for each object pair, with
  $\alpha=\{0.1,0.2,0.5,1\}$. The algorithm stops after 
  $I_{\rm max}=30$ iterations. The performance of QUITE (solid lines) is compared against the BCRB (dashed lines).}\label{fig:figure1}
\end{figure}

\begin{figure}[t]
\centerline{\includegraphics[width=1.00\columnwidth]{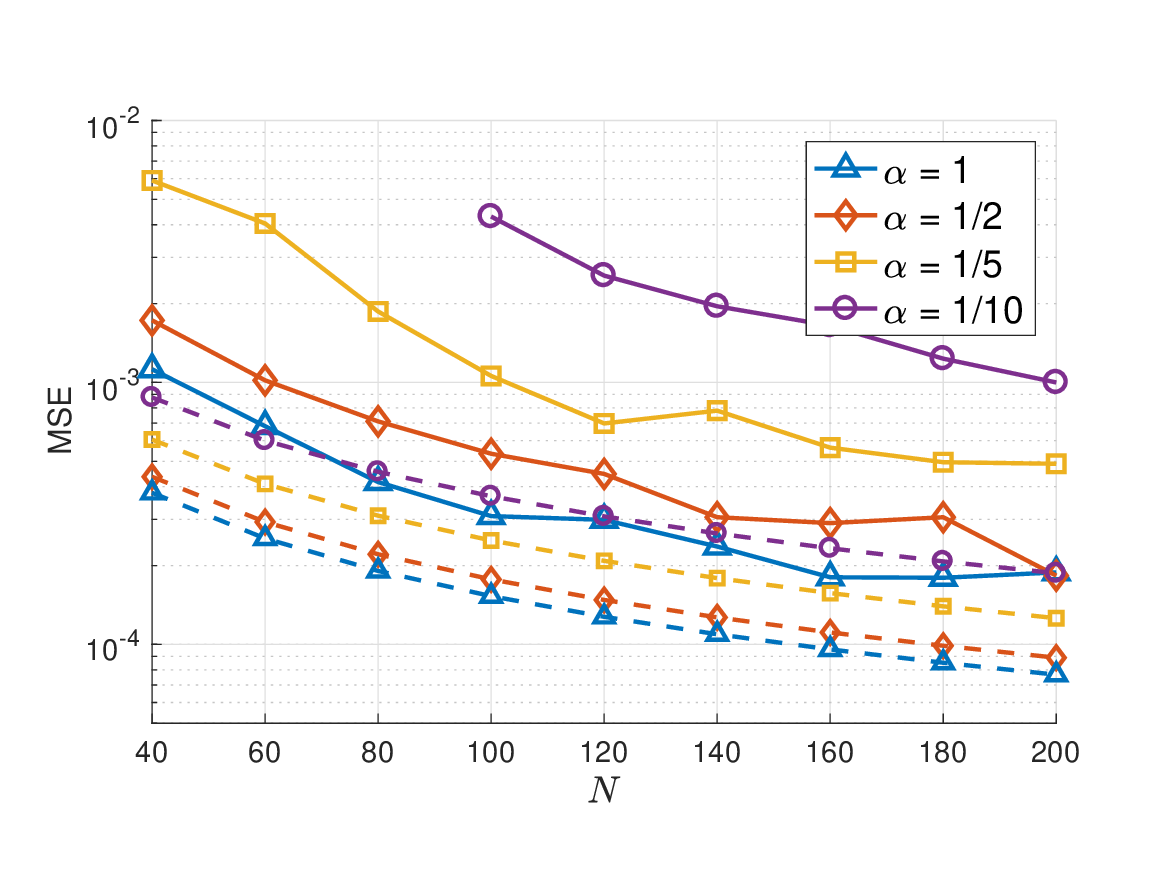}}
\caption{MSE performance of the QUITE algorithm versus the number of
  objects for $K = N$ workers, and Thurstone worker model. The graph is
  regular and has degree $D = 20$. $M = \alpha K$ evaluations
  are carried out for each object pair, with
  $\alpha=\{0.1,0.2,0.5,1\}$. The algorithm stops after 
  $I_{\rm max}=30$ iterations. The performance of QUITE (solid lines) is compared against the BCRB (dashed lines).}\label{fig:figure2}
\end{figure}

In Figures~\ref{fig:figure1} and~\ref{fig:figure2}, we show the MSE
provided by QUITE for several different scenarios. To compute the
MSE, we have adjusted the estimated quality values by performing an
affine transformation, i.e., $\widehat{q}_i \rightarrow A
\widehat{q}_i + B$, where $B$ is the true quality value of the
$N$-th object, which is assumed to be the reference in the LS
algorithm, (i.e., imposing $\widehat{q}_N = 0$), while $A$
is a strictly positive scaling parameter whose value has been obtained through
simulations, as the average over 100 different realizations of the
optimal, MSE-minimizing scaling parameters. Notice that the above transformation does not affect the
final ranking; indeed the rankings induced by the vectors
$\widehat{\qv}$ and $A\widehat{\qv}+B$ are the same, for any $A>0$ and $B$.

Figure~\ref{fig:figure1} shows the MSE performance of QUITE, with the BTL worker model and $I_{\max} = 30$
iterations.  For these experiments, we have not set a stopping threshold $\tau = 0$,  so that QUITE always performs all the $I_{\max}$ iterations.
Curves are parameterized by the value of $\alpha=M/K$, i.e., the number
of evaluations on each edge per worker.  Dashed lines represent the
respective BCRBs. As it can be seen, the MSE decreases with the number
of objects $N$, and with the parameter $\alpha$. This behavior can be
explained as follows.  Since the total number of evaluations is $EM =
10 \alpha N^2$ and the number of parameters to be estimated is $K+N =
2N$, the number of evaluations per unknown parameter is
$\frac{10\alpha}{2} N$, which increases with $N$ and $\alpha$. Hence, the unknown
parameters $\rhov$ and $\qv$ are estimated with increasing
reliability as $N$ increases and as the number of evaluations per edge increases. The ratio between the obtained MSE and
the relative BCRB is always less than one order of magnitude, with a
typical value of 3-5, and slightly improves with $N$.

Figure~\ref{fig:figure2} shows the MSE performance of the QUITE
algorithm, with the Thurstone worker's model. All the other
parameters are same as for Figure~\ref{fig:figure1}. The behavior of
the curves is similar as for the BTL model. However, the Thurstone model shows a slight increase 
in MSE values, especially for low $N$, compared to BTL. 

\subsubsection{Error probability on the estimated ranking}
\begin{figure}[t]
\centerline{\includegraphics[width=1.00\columnwidth]{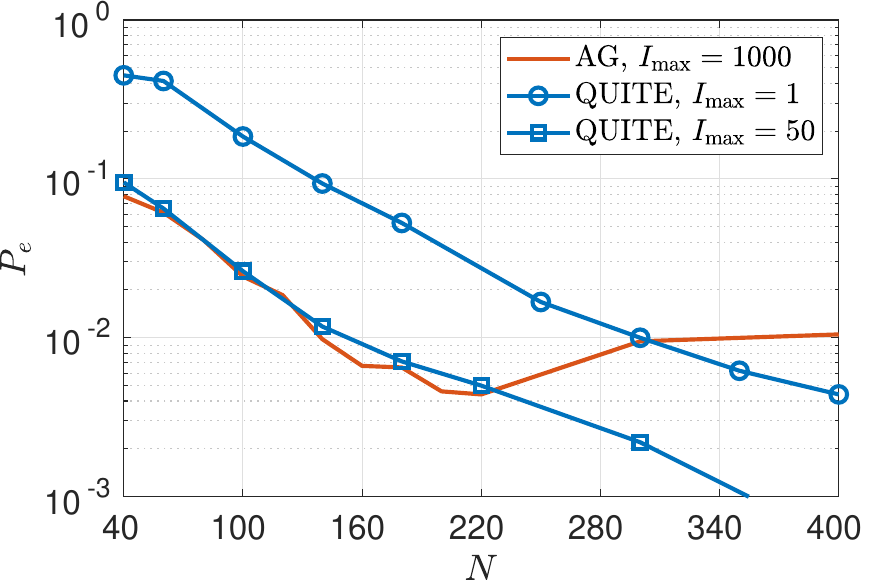}}
\caption{Error probability provided by AG and QUITE algorithms, plotted versus the number
  of objects for $K = N$. Workers obey to the BTL model, with $\alpha = 1/2$ and $\varepsilon$-PAC=0.06.}\label{fig:perr_BTL}
\end{figure}

Now we show the performance of QUITE in terms of the reliability of
the estimated ranking and compare it with the AG algorithm introduced
in~\cite{cinesi} and discussed in Section~\ref{sect:previous_algo}.
For the same scenario considered before, we have performed Monte Carlo
simulations and counted an error whenever the estimated ranking was
not $\epsilon$-quality, as defined in Section~\ref{sec:model}, with
$\epsilon = 0.06$.

For QUITE, we have set a maximum number of iterations equal to
$I_{\max } = 50$ and a stopping threshold $\tau = 10^{-5}$.  For the
AG algorithm, we have performed at most $I_{\max }=1000$ iterations,
with the possibility of stopping iterations in the same way as for
QUITE, and with threshold $\tau = 10^{-5}$, as well. Since the authors in~\cite{cinesi} do not provide details on how to set the
step sizes specified in~\eqref{eq:q_step} and~\eqref{eq:rho_step}, we have set it as
$\lambda_q = \lambda_{\rho} = \frac{N}{5}$, since we have observed that  
scaling with $N$ is beneficial to performance.

\begin{figure}[t]
\centerline{\includegraphics[width=1.00\columnwidth]{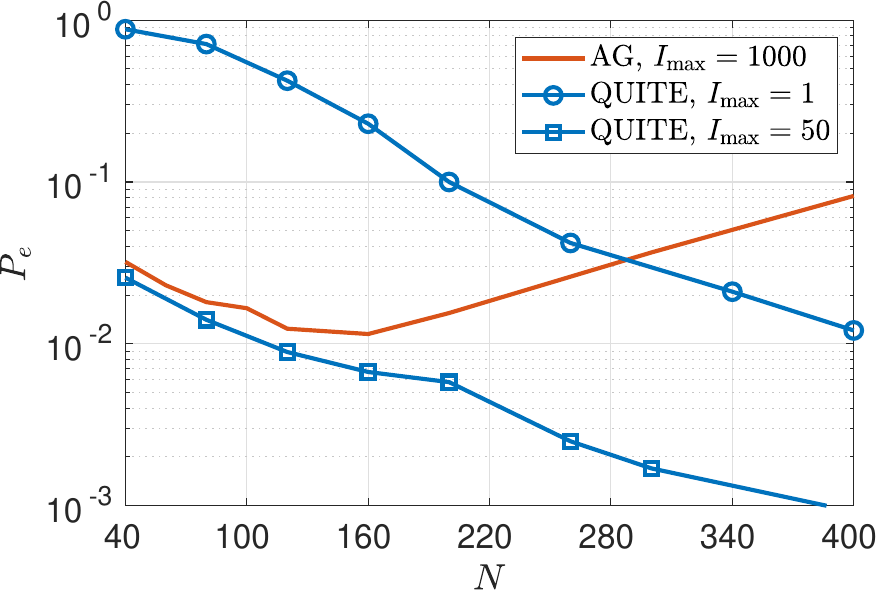}}
\caption{Error probability provided by AG and QUITE algorithms, plotted versus the number
  of objects, for $K = N$. Workers obey to the Thurstone model, with $\alpha = 1/2$ and $\varepsilon$-PAC=0.06.}\label{fig:perr_Thurstone}
\end{figure}

Figures~\ref{fig:perr_BTL} and~\ref{fig:perr_Thurstone} report a
comparison between the performance of QUITE and of AG in the BTL and
Thurstone scenarios, respectively, with $\alpha = \frac1{2}$.  For
QUITE, we also show the performance obtained at the first iteration ($\ell=1$),
i.e., considering the estimates $\widehat{\qv}^{(1)}$ in Algorithm \ref{alg:QUITE-strong}.

First of all, we observe that the
relative performance of QUITE improves with $N$, similarly to what observed for the MSE in Figures~\ref{fig:figure1} and~\ref{fig:figure2}.
In addition to the reasons previously pointed out,  Proposition \ref{prop:theor} also justify
this behavior for $I_{\max}=1$. For small values of $N$, QUITE
performs similarly to AG, with a slight advantage of the latter with
respect to the former under the BTL model and a more significant
advantage of the former with respect to the latter under the Thurstone
model.  However, as $N$ increases (i.e., for $N > 200$), the
relative performance of the AG algorithm
significantly worsens, and for $N>300$ AG is outperformed even by the
first iteration of QUITE.

The same conclusions can be drawn from
Figure~\ref{fig:perr_Thurstone}, where the system parameters of
Figure~\ref{fig:perr_BTL} are employed, but with the Thurstone worker
model instead. In this case, QUITE is never outperformed by the AG
algorithm, while the gap between QUITE and AG increases with $N$
(again, for $N\ge 300$ even the first iteration of QUITE provides
better error probability than AG).

\subsubsection{Impact of the graph structure} 
\begin{figure}[t]
\centerline{\includegraphics[width=1.00\columnwidth]{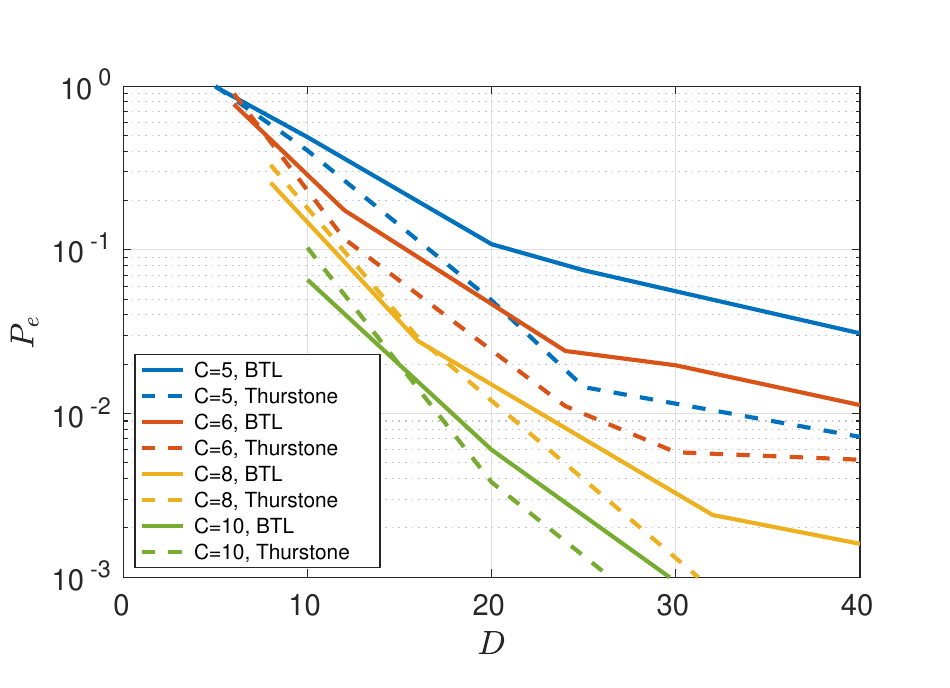}}
\caption{Error probability provided by QUITE versus the graph degree, D, for both the Thurstone and BTL worker models.
The number of objects is $N=200$ and the ranking is $\epsilon$-quality with $\epsilon=0.06$.}\label{fig:figure_degree}
\end{figure}

Finally, Figure~\ref{fig:figure_degree} explores the impact of the
graph structure on QUITE performance.  We set $N=200$
and, to be fair,  we keep fixed the total number of workers'
evaluations, which is equal to $\alpha KND/2=CN^2/2$,  where $C = \alpha D$,  and we vary the
degree $D$ of the regular graph.  
We show performance curves parameterized by different values of $C$.
Note that the choice of the graph has a significant impact on
the QUITE performance.  In particular,  the performance improves
significantly by increasing the graph degree under both the BTL and
Thurstone models. Indeed, the LS algorithm is more efficient in a
graph with larger degree since it can exploits the distance estimates
of a larger number of object pairs to infer the object qualities. Observe also that all curves start at $D=C$ since
$\alpha=C/D$ cannot be larger than 1.

\subsection{Single-stage QUITE versus two-stage}
\begin{figure}[t]
\centerline{\includegraphics[width=1.00\columnwidth]{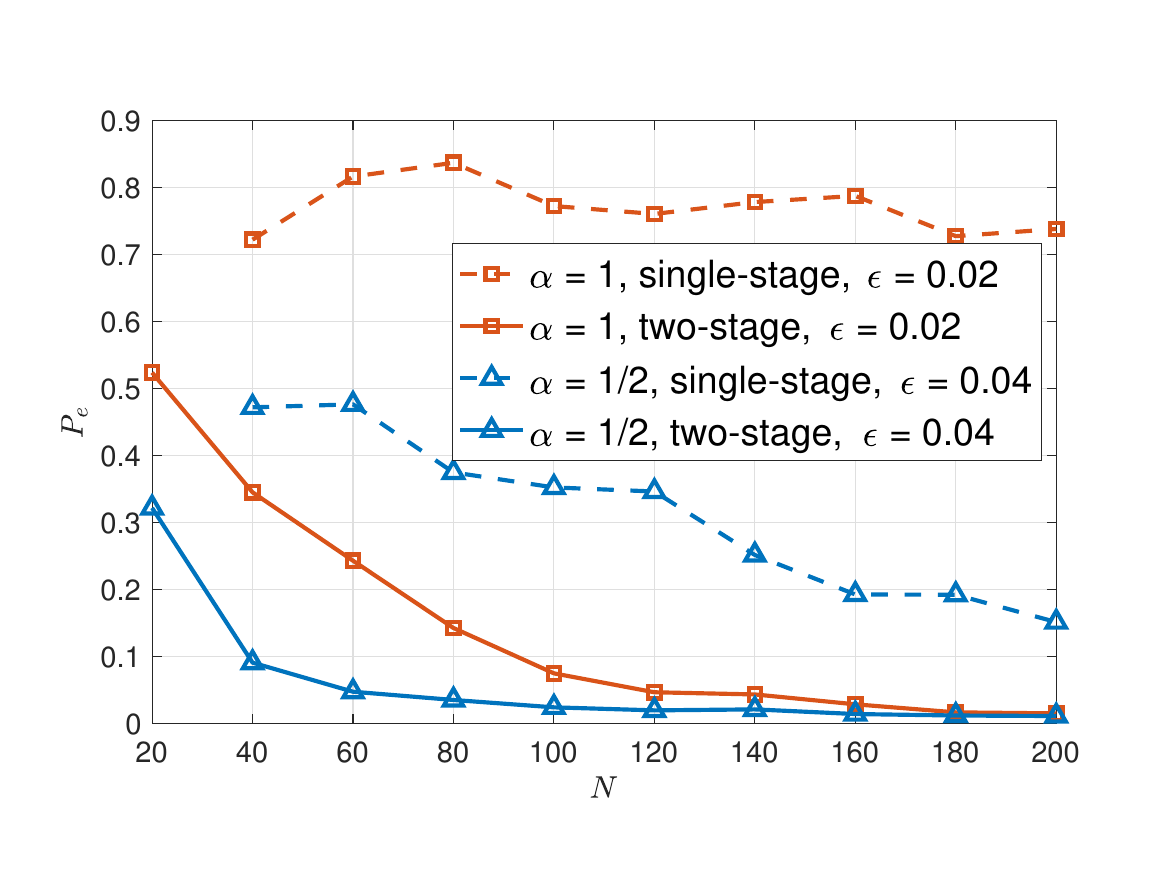}}
\caption{Comparison of single-stage and two-stage QUITE algorithms in the BTL
  scenario by keeping fixed the total number of edges in the graph. In the single-stage case the graph is regular and has degree $D  = 20$.
  In the two-stage case the graph $\Gc_1$ is regular with $D = 10$ and $\Gc_2$ is build according to~\ref{sec:build_G2} with $D=10$. 
  $M = \alpha K$ evaluations are carried out for each object pair,
  with $\alpha=\{0.5,1\}$. }\label{fig:1stage_2stage_tf_12_BTL}
\end{figure}

We now evaluate the performance of the two-stage QUITE algorithm
described in Section~\ref{sec:2stage}, showing the advantage with
respect to single-stage QUITE.  In
Figure~\ref{fig:1stage_2stage_tf_12_BTL}, we consider the BTL model,
with worker reliability drawn from the uniform distribution
$\Uc[1,20]$. The parameters for the first stage are the same as for
Figure \ref{fig:perr_BTL}, except that the graph $\Gc_1$ has degree $D =
10$.  The same graph degree has also been chosen for building the
graph $\Gc_2$, using the procedure detailed in
Section~\ref{sec:build_G2}\footnote{For simplicity, every object pair
  that was already evaluated in the first stage is excluded from
  $\Gc_2$.}. For both stages the number of evaluations per pair is set
to $M = \alpha K$ with $\alpha \in \{0.5,1\}$.
To be fair, in the figure we compare the performance of the two-stage version of QUITE against
a single-stage QUITE with graph degree $D = 20$, so as to keep constant the total number of workers' evaluations.
In counting errors, we have set $\epsilon = 0.04$ for $\alpha = 1/2$ and
$\epsilon = 0.02$ for $\alpha = 1$.

As for assigning workers to pairs in the second stage, following the discussion of Section~\ref{sec:recipe}, we have opted
for the following simple assignment rule. By using the estimates $\widehat{\rhov}$ obtained from the first stage,
we sort the workers in decreasing order of their estimated reliabilities and we partition them into
$1/\alpha$ subsets of $M$ contiguous (in terms of reliability values) workers each. Notice that
$1/\alpha$ is an integer. Then, using $\widehat{\qv}$ obtained from the first stage, we sort the object pairs of $\Gc_2$ in
increasing order of their estimated distances and we partition them
into $1/\alpha$ subsets of contiguous pairs\footnote{Since the number
  of pairs may not be divisible by $1/\alpha$, we may need to add some
  dummy pairs.}. Finally, we assign the $i$-th subset of pairs to the
$i$-th subset of workers, $i = 1,\dots,1/\alpha$. In this way, the
shortest-distance pairs are evaluated by the most reliable workers, as
estimated by the first stage.

From Figure~\ref{fig:1stage_2stage_tf_12_BTL}, it is clear how two-stage
QUITE improves on single-stage QUITE.  This is particularly evident
for $\alpha = 1$, where the single-stage has an error probability
always larger than 0.7, while two-stage QUITE, which relies on the
first stage estimates to build the second stage graph, has a
considerably better performance,  with an error probability below 2\%
for $N = 200$.

\begin{figure}[t]
\centerline{\includegraphics[width=1.00\columnwidth]{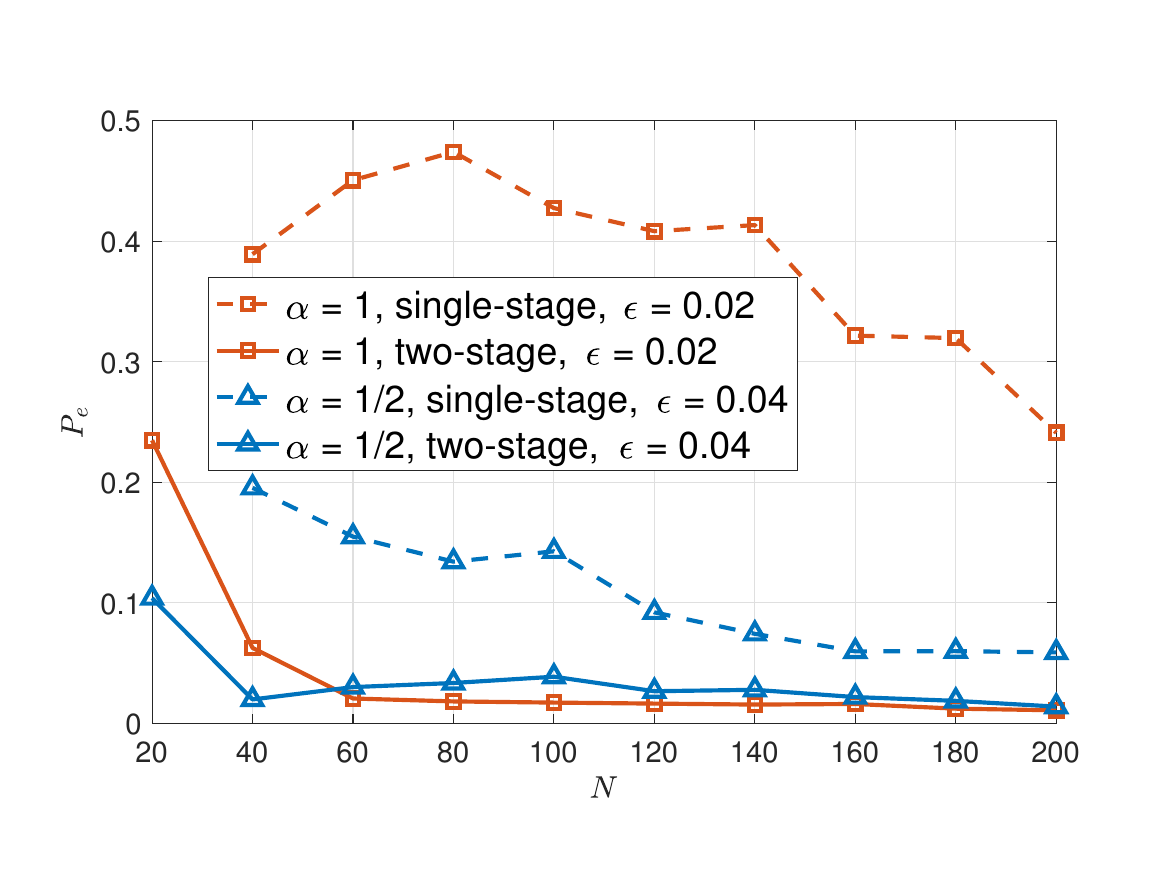}}
\caption{Comparison of single-stage and two-stage QUITE algorithms in the Thurstone
  scenario by keeping fixed the total number of edges in the graph. In the single-stage case the graph is regular and has degree $D  = 20$.
  In the two-stage case the graph $\Gc_1$ is regular with $D = 10$ and $\Gc_2$ is build according to~\ref{sec:build_G2} with $D=10$. 
  $M = \alpha K$ evaluations are carried out for each object pair,
  with $\alpha=\{0.5,1\}$.}\label{fig:1stage_2stage_tf_12_Th}
\end{figure}

Figure~\ref{fig:1stage_2stage_tf_12_Th} shows the comparison between
single-stage and two-stage QUITE in the Thurstone scenario. All the
other parameters are the same as for
Figure~\ref{fig:1stage_2stage_tf_12_BTL}. Also in this case, the error
probability dramatically decreases of even an order of magnitude,
by choosing a two-stage strategy. The advantage of using two-stage
QUITE is especially relevant for the case $\alpha = 1$. For both BTL
and Thurstone scenarios, the gain provided by the second stage seems to
lie in adding new edges between objects with a similar estimated
quality, as estimated by the first stage.

\section{Conclusion\label{sec:conclusion}}

In this paper, we have faced with the problem of ranking a set of
objects by means of the evaluations provided by a pool of heterogeneous
workers.  In particular, we have answered in the affirmative the
following question: ``Does an estimate of the workers' reliabilities
improve the ranking?''  Not only: we have proposed QUITE, an iterative
algorithm that comes into two flavors: single-stage or two-stage.
Even single-stage QUITE improves on algorithms from the literature.
Two-stage QUITE capitalizes on the first stage estimates to ask
further evaluations to the same pool of workers, yielding a dramatic
improvement, in terms of ranking correctness.

\appendices
\section{Proof of Proposition~\ref{prop:BCRB}\label{app:proof_BCRB}}
We start by computing the BIM in~\eqref{eq:info_matrix}. We first note
that $\log \PP(\Wc,\thetav) =\log\PP(\Wc | \thetav)+\log f_{\thetav}(\thetav)$
where $\log f_{\thetav}(\thetav)$ can be further expanded as
$\log f_{\rhov}(\rhov)+\log f_{\qv}(\qv)$.  Therefore
\begin{eqnarray}
  \Mm &=& -\EE_{\Wc,\thetav}\left[\nabla_{\thetav}\nabla_{\thetav}\Tran\log\PP(\Wc|\thetav)\right]\non
          &&\quad -\EE_{\thetav}\left[\nabla_{\thetav}\nabla_{\thetav}\Tran\log f_{\qv}(\qv)\right]{-}\EE_{\thetav}\left[\nabla_{\thetav}\nabla_{\thetav}\Tran\log f_{\rhov}(\rhov)\right] \non
  &=& -\EE_{\Wc,\thetav}\left[\left[\begin{array}{cc}\nabla_{\qv}\nabla_{\qv}\Tran & \nabla_{\qv}\nabla_{\rhov}\Tran\\
\nabla_{\rhov}\nabla_{\qv}\Tran & \nabla_{\rhov}\nabla_{\rhov}\Tran\end{array}\right]
                                  \log\PP(\Wc|\thetav) \right]\non
  && \quad-\EE_{\thetav}\left[\begin{array}{cc}\nabla_{\qv}\nabla_{\qv}\Tran\log f_{\qv}(\qv) & \zerov \\
                                                                                                                      \zerov & \nabla_{\rhov}\nabla_{\rhov}\Tran\log f_{\rhov}(\rhov) \end{array}\right]\non
\end{eqnarray}
Since the random variables $q_i$'s are assumed i.i.d., by
recalling~\eqref{eq:p_rho_p_q}, we have
$-\nabla_{\qv}\nabla_{\qv}\Tran\log f_{\qv}(\qv)=\beta_q\Id$ where
$\beta_q = -\EE_q\left[\frac{\dd^2\log f_q(q)}{\dd q^2}\right]$.
Similarly, $-\nabla_{\rhov}\nabla_{\rhov}\log f_{\rhov}(\rhov)=\beta_{\rho}\Id$ where
$\beta_{\rho} = -\EE_{\rho}\left[\frac{\dd^2\log f_\rho(\rho)}{\dd
    \rho^2}\right]$.

Next, we consider the term
$-\EE_{\Wc,\thetav}\left[\nabla_{\qv}\nabla_{\qv}\Tran\log\PP(\Wc|\thetav)\right]$
where the probability $\PP(\Wc|\thetav)=\PP(\Wc|\rhov,\qv)$ can be expanded as
\begin{equation}\label{eq:P_expand1}
  \log\PP(\Wc|\thetav) = \sum_{e=1}^E \sum_{k\in \Kc_e}\log
  F(x_{e,k})
\end{equation}
or equivalently as
\begin{equation}\label{eq:P_expand2}
  \log\PP(\Wc |\thetav) = \sum_{k=1}^K \sum_{e\in \Ec_k}\log
  F(x_{e,k})
\end{equation}
since the workers' answers are independent. Note that such expressions are in fact functions of the
distances $\dv$ which are related to the object qualities $\qv$
through~\eqref{eq:d_versus_q}. Then $\PP(\Wc|\rhov,\qv)=\PP(\Wc|\rhov, \dv)=\PP(\Wc|\rhov, \Gammam\Tran\qv)$
and
\begin{eqnarray}
  \nabla_{\qv}\left[\log \PP(\Wc|\thetav)\right]  &=& \nabla_{\qv}\left[\log\PP(\Wc|\rhov, \Gammam\Tran\qv)\right] \non
  &=& \Gammam \nabla_{\dv}\left[\log\PP(\Wc|\rhov, \dv)\right]
\end{eqnarray}
where we applied the derivative chain rule. It immediately follows
\begin{eqnarray}
 &&\hspace{-10ex} -\EE_{\Wc,\thetav}\left[\nabla_{\qv} \nabla_{\qv}\Tran\log \PP(\Wc|\thetav) \right] \non
  &=& -\Gammam \EE_{\Wc,\thetav}\left[\nabla_{\dv} \nabla_{\dv}\Tran\log
      \PP(\Wc|\rhov,\dv)\right]\Gammam\Tran\non
  &=& \Gammam \Deltam_q\Gammam\Tran
\end{eqnarray}
where $\Deltam_q = -\EE_{\Wc,\thetav}[\nabla_{\dv} \nabla_{\dv}\Tran\log
  \PP(\Wc|\rhov,\dv)]$.
Now, let us define $F_{e,k}\triangleq F(x_{e,k})$, $F'_{e,k}\triangleq F'(x_{e,k})$ and $F''_{e,k}\triangleq F''(x_{e,k})$
where $F'$ and $F''$ are the first and second derivatives of $F$, respectively.
Then, by using~\eqref{eq:P_expand1}
\begin{eqnarray}
[\Deltam_q]_{e,e'}&\hspace{-2ex}=& \hspace{-2ex}-\EE_{\Wc,\thetav}\left[\frac{\partial^2\log \PP(\Wc|\rhov,\dv) }{\partial d_e\partial d_{e'}}\right] \non
                                        &\hspace{-2ex}=&\hspace{-2ex} \left\{
      \begin{array}{ll} 0 & \mbox{ if } e{\neq} e' \\
      -\EE_{\thetav}\left[\displaystyle\sum_{k\in \Kc_e}\rho_k^2\EE_{w_{e,k}}\left[\widetilde{F}_{e,k}\right]\right]
                          & \mbox{ if } e{=}e' \end{array}\right.\label{eq:A}     
\end{eqnarray}
where
$\widetilde{F}_{e,k}\triangleq\frac{F_{e,k}''F_{e,k}-(F'_{e,k})^2}{(F_{e,k})^2}$
and, in the derivation, we used the identity $(1-2w_{e,k})^2=1$ which
holds for all $k$ and $e$, since $w_{e,k}\in\{0,1\}$.  Now, we exploit
the property $F(-x) = 1-F(x)$ of the function $F$ which implies
$F'(-x) = F'(x)$ and $F''(-x)=-F''(x)$.  Such relations can be used to
simplify the average in~\eqref{eq:A}. Indeed, we obtain
\begin{eqnarray}
  &&\hspace{-10ex}-\EE_{w_{e,k}}\left[\widetilde{F}_{e,k}\right]\non
  &=&{-}\widetilde{F}_{e,k,0}\PP(w_{e,k}{=}0{|}\rho_k{,}d_e){-}\widetilde{F}_{e,k,1}\PP(w_{e,h}{=}1{|}\rho_k{,}d_e) \non
  &=&\frac{(F'(\rho_kd_e))^2}{F(\rho_kd_e)(1-F(\rho_kd_e))}
\end{eqnarray}
where $\widetilde{F}_{e,k,0}\triangleq \widetilde{F}_{e,k}\big|_{w_{e,k}{=}0}$ and $\widetilde{F}_{e,k,1}\triangleq \widetilde{F}_{e,k}\big|_{w_{e,k}{=}1}$.
It follows that $\Deltam_q$ is a diagonal matrix whose elements are given by
\begin{eqnarray}
[\Deltam_q]_{e,e'} &=& \left\{
      \begin{array}{ll} 0 & \mbox{ if } e\neq e' \\
      |\Kc_e|\EE_{\rho,d}\left[\frac{\rho^2(F'(\rho d))^2}{F(\rho d)(1-F(\rho d))}\right]
                          & \mbox{ if } e=e' \end{array}\right.\label{eq:A1}      
\end{eqnarray}
since the distances $d_e$, $e=1,\ldots,E$ are identically distributed and the parameters $\rho_k$, $k=1,\ldots,K$ are i.i.d..

Similarly, by using~\eqref{eq:P_expand2}, the element $(k,k')$ of the matrix $\Deltam_{\rho}=-\EE_{\Wc,\thetav}\left[\nabla_{\rhov}\nabla_{\rhov}\Tran\log\PP(\Wc|\thetav)\right]$ can be written as
\begin{eqnarray} [\Deltam_{\rho}]_{k,k'}
  &=& -\EE_{\Wc,\thetav}\left[\frac{\partial^2}{\partial \rho_k\partial \rho_{k'}}\log\PP(\Wc|\thetav)\right]\non
  &=&\left\{\begin{array}{ll} 0 & \mbox{ if } k\neq k' \\  |\Ec_k| \EE_{\rho,d}\left[\frac{d^2(F'(\rho d))^2}{F(\rho d)(1-F(\rho d))}\right] & \mbox{ if } k=k' \end{array}\right. \label{eq:B}
\end{eqnarray}
Finally, we consider the matrix $-\EE_{\Wc,\thetav}\left[\nabla_{\qv}\nabla_{\rhov}\Tran\log\PP(\Wc|\thetav)\right]$ whose $(e,k)$-th element is given by 
\begin{eqnarray}
  &&\hspace{-10ex}-\EE_{\Wc,\thetav}\left[\nabla_{\qv}\nabla_{\rhov}\Tran\log\PP(\Wc|\thetav)\right]_{e,k} \non
  &=& -\EE_{\Wc,\thetav}\left[\frac{\partial^2}{\partial \rho_k\partial d_{e}}\log\PP(\Wc|\thetav)\right]\non
  &=& \left\{ \begin{array}{ll} 0 & \mbox{ if } k \notin \Kc_e \\ \EE_{\thetav}\left[\frac{\rho_kd_e(F'(\rho_kd_e))^2}{(F(\rho_kd_e)(1-F(\rho_kd_e))}\right]& \mbox{ if } k \in \Kc_e \end{array} \right.\label{eq:C} 
\end{eqnarray}
Note that
$\frac{\rho_kd_e(F'(\rho_kd_e))^2}{(F(\rho_kd_e)(1-F(\rho_kd_e))}$ is
an odd function of $d_e$. Therefore, when averaged over $d_e$, whose
density is even, (recall that $d_e$ is the difference of two i.i.d. distributions)
it provides zero. So the term in~\eqref{eq:C} is always zero and the BIM
takes the expression in~\eqref{eq:BIM}.

\section{extending the proofs of  \cite{noiTNSE}\label{extension-proofs}}
In \cite{noiTNSE} it is assumed that workers have the same
reliability, $\rho$, which is perfectly known. Therefore, as initial
estimate of distance $d_e$ it is set: \beq
\widehat{d}_e=\frac{1}{\rho} F^{-1}(\widehat{p}_e) \qquad \text{where}
\qquad \widehat{p}_e=1-|\Kc_e |^{-1}\sum_{ k\in \Kc_e}w_{e,k}.  \eeq
Now, the estimate $\widehat{p}_e$, of the probability $p_e$ can be
written as $\widehat{p}_e= {p}_e+y_e$ where $y_e$ is the estimation
error. Since the estimate $\widehat{p}_e$ is unbiased we can also
write $p_e= \mathbb{E}[ \widehat{p}_e] =F(d_e\rho)$.  The estimation
error on the distance $d_e$ is related to $y_e$ as follows:
\beq \label{old-asy} z_{e} = \widehat{d}_{e} - d_{e} =\frac{1}{\rho}
\left.\frac{\mathrm{d} F^{-1}(p)} {\mathrm{d} p}\right|_{p=p_{e}}
y_{e}+O\left(y_{e}^2 \right)\,.  \eeq where by construction $\inf_e
\left. \frac{\mathrm{d} F^{-1}(p)} {\mathrm{d} p}\right|_{p=p_{e}}>0$.

In the heterogeneous case, where reliability  of individual workers is not known  and distribution 
$f_\rho(\rho) $ is given,  we can replace the previous distance estimate with: 
\beq \label{eq:init_est_dist-1}
\widehat{d}_e=G^{-1}(\widehat{p}_e) \qquad \text{where} \qquad G(d)=\int F(\rho d)f_\rho(\rho) \dd \rho.
\eeq
$\widehat{p}_e=  {p}_e+y_e$ and $p_e= \mathbb{E}[\widehat{p}_e]=G(d_e)$. Therefore, it turns out that: 
\beq  \label{new-asy}
z_e = \widehat{d}_e - d_e = \left.\frac{\mathrm{d} G^{-1}(p)} {\mathrm{d} p}\right|_{p=p_{e}} y_{e}+O\left(y_{e}^2 \right)\,.
\eeq
Now, observe that, as long as $\inf_e \left. \frac{\mathrm{d} G^{-1}(p)} {\mathrm{d} p} \right|_{p=p_{e}} >0$,   
 \eqref{old-asy} and \eqref{new-asy} have exactly the same structure.
Therefore Proposition 6.1 in\cite{noiTNSE} (reported  below)   extends rather easily to our case.
 \begin{prop}\label{prop-YZ} (\cite[Proposition 6.1]{noiTNSE})
 	For any $ \epsilon>0$ and $\delta>0$, there exists $\beta(\epsilon,\delta)$ such that, as $N\to \infty$, 		
 	\begin{equation}\label{eq:prop-YZ}
 		\PP\left(\sup_{e \in \Ec}|y_{e}|>\epsilon\right) \mathord{<} \delta  \mbox{ and } \PP\left(\sup_{e  \in \Ec}|z_{e}|>\epsilon\right) < \delta
 	\end{equation}
 	provided that for every edge $e\in \mathcal{E}$ we have $|\mathcal{K}_e| > \beta(\epsilon, \delta) \log N $   with $ \beta(\epsilon, \delta)= O\left(\frac{1}{\epsilon^2} \frac{ \log \frac{N}{\delta} } {\log N} \right) $
 	and the total number of pairs  is $|\Ec| =O(N)$.
 \end{prop}

\pf
We first use the union bound and write $\PP\left(\sup_{e \in \Ec}|y_{e}|>\epsilon\right)\le \sum_{e\in \Ec}\PP(y_{e}>\epsilon)+\sum_{e\in \Ec}\PP(-y_{e}>\epsilon)$.
We then observe that the moment generating function (MGF) $\phi_{y_{e}}(t)$ of $y_{e}$ is given by:
\[
\phi_{y_e}(t){=} \left(\int_\rho  \ee^{-\frac{tF(\rho d_e)}{|\mathcal{K}_e|}} (1{+}  F(\rho d_e) (\ee^{\frac{t}{|\mathcal{K}_e|}}{-}1))f_\rho(\rho) \mathrm{d} \rho \right)^{|\mathcal{K}_e|}\,. 
\]
By applying the mean-value  theorem,   there exists a $\rho^*\in[\rho_{\min}, \rho_{\max}]$ such that 
\[
\phi_{y_e}(t){=} \left( \ee^{-\frac{tF(\rho^*d_e)}{|\mathcal{K}_e|}} (1+  F(\rho^*d_e) (\ee^{\frac{t}{|\mathcal{K}_e|}}-1)) \right)^{|\mathcal{K}_e|}.
\]
Next, we bound  $\PP(y_{i,j}>\epsilon)$ by applying  the Chernoff bound:
\[
\PP(y_{e}>\epsilon) \le \inf_{t>0} \frac{\phi_{y_e}(t)}{\ee^{\epsilon t}} \le  \frac{\phi_{y_e}(t)}{\ee^{\epsilon t}}\,.
\]
By setting $t=\zeta \log N $, and  $W_e=\beta \log N$, for a sufficiently large $\beta=\beta(\epsilon, \delta)$ ,
we have 
\[
\PP(y_{e}{>}\epsilon) {\le} \ee^{\left(\beta \log(  1{+} F(\rho^*d_e)(\ee^{\frac{\zeta}{\beta}}{-}1){-}\zeta F(\rho^*d_e)\right) \log N},
\]
with $ \beta \log( 1+ F(d_e\rho^*)(\ee^{\frac{\zeta}{\beta}}-1))=
\beta ( \log (1 +F(d_e\rho^*)\frac{\zeta}{\beta}+
O(\frac{\zeta^2}{\beta^2}) ) =\zeta F(d_e\rho^*)+
O(\frac{\zeta^2}{\beta})$.  Now, for $\beta$ sufficiently large, we
can always assume that the above error term (i.e. the term $
O(\frac{\zeta^2}{\beta})$) can be made smaller than
$\frac{\epsilon\zeta}{2}$ and therefore
$\PP(y_{i,j}>\epsilon)<N^{\frac{\zeta\epsilon}{2}}$, with
$\zeta\frac{\epsilon}{2}>1$.  This implies $\sum_{e\in \Ec}
\PP(y_{e}>\epsilon) \le N^{1-\frac{\epsilon\zeta}{2}}\to 0$ as $N\to
\infty$. As a consequence, the statement has been proved for $\delta$
bounded away from 0, since, as a result of the previous relationships,  we can
choose $\beta= O(\frac{1}{\epsilon^2})$.  Finally,  for $\delta= o(1)$,
by imposing that $ N^{1-\frac{\epsilon\zeta}{2}}>\delta$, we get that
$ \beta(\epsilon, \delta)=
O\left(\frac{1}{\epsilon^2}\frac{\log\frac{N}{\delta}}{\log N}
\right)$ for the more general case.  Similarly, the term
$\sum_{e\in \Ec}\PP(-y_{e}>\epsilon)$ also tends to 0 as $N$ grows.

As for the second claim of the proposition we can write again
$\PP\left(\sup_{e \in \Ec}|z_{e}|>\epsilon'\right)\le \sum_{e\in \Ec}\PP(z_{e}>\epsilon)+\sum_{e\in \Ec}\PP(-z_{e}>\epsilon)$.
We then recall that $z_{e}=\widehat{d}_{e}-d_{e}$, $\widehat{d}_{i,j}=G^{-1}(y_{e}+p_{e})$, and $d_{e}=q_i-q_j$.
It follows that 
\begin{eqnarray}
	\PP(z_{i,j}>\epsilon') &=&   \PP\left(G^{-1}(y_{i,j}+p_{i,j}) -(q_i-q_j)>\epsilon'\right) \non
	&=&\PP\left(G^{-1}(y_{i,j}+G(q_i-q_j)) >\epsilon'+q_i-q_j\right)\non
	&=& \PP\left(y_{i,j}+G(q_i-q_j) >G(\epsilon'+q_i-q_j)\right)\non
	&=& \PP\left(y_{i,j} \mathord{>}G(\epsilon'+q_i-q_j)- G(q_i-q_j)\right)\,. 
\end{eqnarray}
By defining $\epsilon\triangleq G(\epsilon'+F(q_i-q_j))- G(q_i-q_j)>0$,  
the convergence of $\PP(z_{e}>\epsilon')$ to 0 as $N$ grows
immediately follows.  Similarly, it is straightforward to prove the
convergence to 0 of the term $\sum_{e \in \Ec}\PP(-z_{e}>\epsilon)$.
Then,  the main results of  \cite{noiTNSE}   (in particular Propositions 4.1, 4.2, 4.3  4.4 and 4.5)
	 can be immediately extended to our  case.

\bibliographystyle{IEEEtran}
\bibliography{refs_multiclass}




\end{document}